\documentclass[10pt,twocolumn,letterpaper]{article}

\usepackage{cvpr}              

\usepackage{bbm}
\usepackage{graphicx}
\usepackage{amsmath}
\usepackage{amssymb}
\usepackage{amsthm}
\usepackage{mathrsfs}
\usepackage{amsfonts}
\usepackage{booktabs}
\usepackage{multirow}
\usepackage{multicol}
\usepackage[table]{xcolor}
\usepackage{color, colortbl}
\usepackage{pifont}

\usepackage[pagebackref,breaklinks,colorlinks]{hyperref}

\usepackage[capitalize]{cleveref}
\crefname{section}{Sec.}{Secs.}
\Crefname{section}{Section}{Sections}
\Crefname{table}{Table}{Tables}
\crefname{table}{Tab.}{Tabs.}

\def\eg{\emph{e.g.}} 
\def\ie{\emph{i.e.}} 

\def\etc{\emph{etc.}}
\def\etal{\emph{et al.}}

\newcommand{\thename}{TinyDet}

\begin{document}
\title{\thename: Accurate Small Object Detection in Lightweight Generic Detectors$^\star$}
\author{Shaoyu Chen\textsuperscript{1},  Tianheng Cheng\textsuperscript{1}, 
Jiemin Fang\textsuperscript{2}, Qian Zhang\textsuperscript{3},  Yuan Li\textsuperscript{4},  Wenyu Liu\textsuperscript{1},   Xinggang Wang\textsuperscript{1,$\dagger$}\\
[2mm]
  \textsuperscript{1} School of EIC, Huazhong University of Science and Technology \\
  \textsuperscript{2} Institute of AI, Huazhong University of Science and Technology \\
  \textsuperscript{3} Horizon Robotics \quad
  \textsuperscript{4} Google \\
[2mm]
Source code and pretrained models are available at: \href{https://github.com/hustvl/TinyDet}{hustvl/TinyDet}
}
\maketitle

\let\thefootnote\relax\footnote{$^\star$ A letter version of this paper is published in SCIENCE CHINA Information Sciences (SCIS) with \href{https://doi.org/10.1007/s11432-021-3504-4}{doi.org/10.1007/s11432-021-3504-4}. Please cite the SCIS version. $^\dagger$ Corresponding author: Xinggang Wang (\texttt{xgwang@hust.edu.cn}).}

\begin{abstract}
Small object detection requires the detection head to scan a large number of positions on image feature maps, which is extremely hard for computation- and energy-efficient lightweight generic detectors. To accurately detect small objects with limited computation, we propose a two-stage lightweight detection framework with extremely low computation complexity, termed as \thename. It enables high-resolution feature maps for dense anchoring to better cover small objects, proposes a sparsely-connected convolution for computation reduction, enhances the early stage features in the backbone, and addresses the feature misalignment problem for accurate small object detection. On the COCO benchmark, our \thename-M achieves $30.3$ AP and $13.5$ AP$^{s}$ with only $991$ MFLOPs, which is the first detector that has an AP over $30$ with less than $1$ GFLOPs; besides, \thename-S and \thename-L achieve promising performance under different computation limitation. 
\end{abstract}

\begin{figure}[!t]
    \centering
    \includegraphics[width=\linewidth]{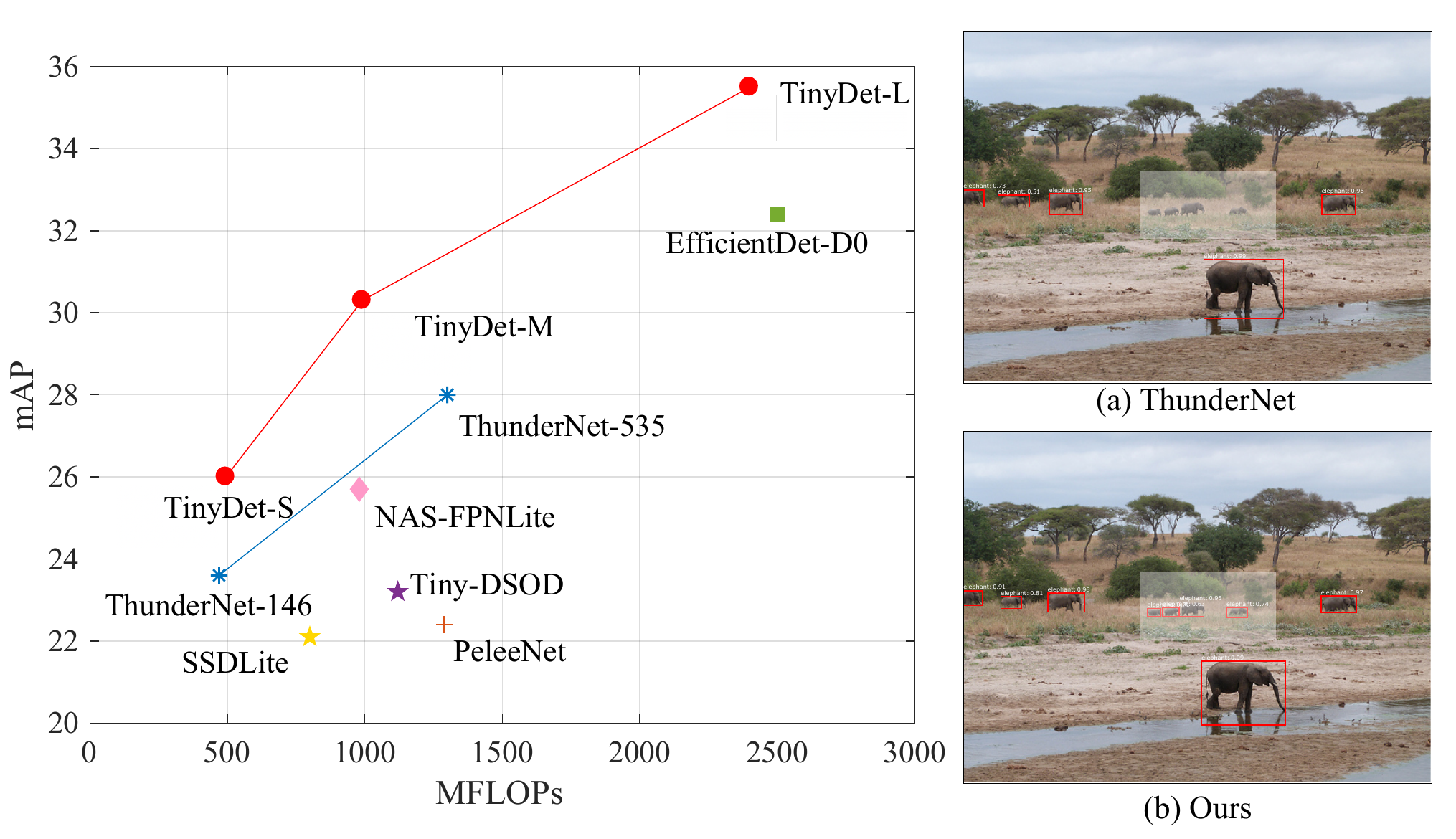}
    \caption{Model FLOPs \textit{vs} COCO Accuracy. Our \thename\ achieves higher mAP with less computation cost compared with other detectors.}
    \label{fig:leading}
\end{figure}

\section{Introduction}
Object detection plays an important role in computer vision, and gradually becomes the technical foundation of many applications, such as autonomous driving, remote sensing and video surveillance. However, the inference procedure of advanced detection models costs massive computational resources, which makes them hard to be applied on resource-constrained mobile or edge devices. To widely apply artificial intelligence (AI), ``Tiny AI" models that are both computation-efficient and energy-efficient are becoming more and more popular. In this paper, we design a lightweight generic object detection framework, termed as \thename, for efficient and accurate object detection, especially small objects, with low computation cost.

In recent years, many innovative and representative lightweight detection models \cite{SandlerHZZC18,WangBL18,QinLZBYPS19,ChenLMXJ19,TanPL19} have been proposed for the better trade-off between the computation cost and accuracy. To reduce the computation cost, they usually downscale the input image by a large ratio and perform object detection on small feature maps. For example, Pelee \cite{WangBL18} takes $304\times304$ input and the largest feature map used for detection in Pelee is $19\times19$. ThunderNet \cite{QinLZBYPS19} takes $320\times320$ input, and only uses a single feature map with the resolution of $20\times20$ for detection. For comparison, large models like Faster R-CNN \cite{RenHGS15} with feature pyramid network (FPN) \cite{LinDGHHB17} takes $800\times1333$ input and the largest feature map is as large as $200\times333$. Performing object detection with small input images and on small feature maps is helpful to reducing computation cost. 
However, small feature maps have no detailed information and poor positional resolution. Previous lightweight detectors have very limited ability to detect small objects. They sacrifice the detection performance of small objects for high efficiency.

The capacity of detecting small objects is of great importance for object detection based applications and is a key factor for evaluating an object detection model. As demonstrated in the COCO dataset, approximately $41\%$ objects are small (area $< 32^2$)~\cite{LinMBHPRDZ14}. In this paper, we target at boosting the small object detection performance in lightweight generic detection networks. Based on the good practices of designing lightweight networks, we propose \thename, which is a two-stage detector with high-resolution (HR) feature maps for dense anchoring. HR feature maps significantly improve the small object detection ability but also bring much more computation cost. To release the contradiction between resolution and computation, we propose TinyFPN and TinyRPN by introducing a sparsely-connected convolution (SCConv), which keeps both high resolution and low computation.
We improve the backbone network for better small object detection performance. Small object detection relies more on detailed information in shallow features. We keep more detailed information by allocating more computation to the early stages.
Besides, we observe that severe feature misalignment exists in lightweight detectors. The feature misalignment accumulates layer by layer and is passed to the detection part, affecting the precision of regression in both RPN and R-CNN head. Small objects are much more sensitive to such positional misalignment. By eliminating the misalignment, the detection performance of small objects is significantly improved.

Our contributions can be summarized as:
\begin{itemize}
    \item In lightweight detection networks, for the first time, we enable the high-resolution detection feature map (\ie, $80\times80$) for dense anchoring, which is essential for detecting small objects.
    \item We propose TinyFPN and TinyRPN with the sparsely-connected convolution to perform efficient object detection with high-resolution detection feature maps.
    \item By enhancing the early stages in the backbone and addressing the misalignment problem in \thename, we further improve the detection results of small objects.
    \item Our \thename\ models have strong performance and little computation budget on the COCO \textit{test-dev2017} set as shown in Fig.~\ref{fig:leading}. \thename-M achieves 30.3 mAP with only 991 MFLOPs, which is the state-of-the-art result among lightweight detectors. Notably, the small object detection performance is outstanding, the APs of \thename-S and \thename-M are two times of that of ThunderNets.
\end{itemize}

\section{Related Work}
\paragraph{Lightweight Object Detector}
Lightweight models for generic object detection has witnessed rapid developments in recent years. Firstly, advances of lightweight classification network design methods~\cite{HowardZCKWWAA17,SandlerHZZC18,HowardSCCCTWZPVLA19,MaZZS18,ZhangZLS18} directly boost the development of lightweight object detection. The lightweight classification networks are often directly adopted as backbones of detectors to extract features, \eg, the hand-crafted MobileNetV2~\cite{SandlerHZZC18} is used in SSDLite~\cite{SandlerHZZC18} and the neural architecture searched EfficientNet~\cite{TanL19} is used in EfficientDet~\cite{TanPL19}. But detection and classification need different backbones~\cite{Li_2018_ECCV}. To better match characteristics of detection tasks, in many lightweight detectors~\cite{QinLZBYPS19,ChenLMXJ19}, specialized backbones are proposed, based on existing classification networks.
Secondly, well developed pipelines~\cite{LiuAESRFB16,RenHGS15,LinDGHHB17,LinGGHD17} for object detection also build a solid foundation for lightweight detector research.
Most lightweight detectors~\cite{RedmonF18,SandlerHZZC18,WangBL18,TanPL19} follow the compact one-stage architecture. PeleeNet~\cite{WangBL18} is only built with conventional convolutions without using the popular mobile convolution. RefineDetLite~\cite{ChenLMXJ19} makes design specialized for CPU-only devices. EfficientDet~\cite{TanPL19} proposes a weighted bi-directional feature pyramid network for easy and fast feature fusion and builds a scalable detection architecture across a wide spectrum of resource constraints. Two-stage detectors, with more complicated pipelines, are usually thought to be more time consuming in inference phase. However, some~\cite{DaiLHS16,LiPYZDS17,QinLZBYPS19} prove that two-stage detectors can also be as efficient as one-stage ones if the second stage is made lightweight enough. The two-stage paradigm tends to be better performing at detecting small objects. Thus, we follow the two-stage paradigm to design our detector by considering both efficiency and accuracy.

\paragraph{Small Object Detection}
Detecting small objects from the video and image has a high-profile in computer vision, remote sensing, autonomous driving, \etc. Liu~\etal~\cite{LiuAESRFB16} creates more small object training examples by reducing the size of large objects. D-SSD~\cite{FuLRTB17}, C-SSD~\cite{XiangZYA18}  F-SSD~\cite{CaoXYLSW2017} and ION~\cite{BellZBG16} focus on building appropriate context features for small object detection. Hu~\etal~\cite{HuR17} makes use of a coarse image pyramid and uses two times upsampled input images to detect small faces.  Several studies, such as \cite{BaiZDG18,LiLWXFY17,NohBLSK19}, use generative adversarial network (GAN)~\cite{GoodfellowPMXWOCB14} to generate super-resolved features for small object detection. Larger input resolution and super-resolution methods bring much more computation cost and are not suitable for lightweight detector design. Our work is slightly related to high-resolution network (HRNet)~\cite{wang2019deep}, which maintains a high-resolution representation in the whole network for position-sensitive visual recognition tasks. We find that there are no papers that focus on detecting small objects in lightweight generic detectors. In this paper, we target at accurate small object detection while keeping low computation budget. 

\paragraph{Sparse Convolution}
Reducing the redundancy in deep convolutional neural networks is an important direction to explore when designing lightweight detection networks. Many previous works dedicate to reduce the redundancy by making the connections in convolution sparser. Liu~\etal~\cite{LiuWFTP15} adopts sparse decompositions to get sparse convolutional networks and lower the computational complexity. 
The depth-wise separable convolution~\cite{HowardZCKWWAA17,Chollet17}, a kind of factorized convolution, is widely adopted for reducing computation and model size. And the group convolution is used in \cite{ZhangZLS18} for efficient model design. Our work combines the depth-wise convolution and group convolution in a novel form to build extremely lightweight FPN and RPN.

\begin{table*}[!t]
    \centering
    \caption{Specification for the backbone of \thename-M. Bneck denotes the inverted residual bottleneck structure~\cite{SandlerHZZC18}. ExSize denotes expansion size. SE denotes the squeeze-and-excitation module~\cite{hu2018squeeze}. NL denotes the type of the used nonlinearity. HS denotes h-swish~\cite{HowardSCCCTWZPVLA19} and RE denotes ReLU. FPN denotes whether the output of the block is fed into FPN. Enhancement denotes modifications compared with the original MobileNetV3 and "c" denotes channel. }
    \label{tab:backbone}
    \resizebox{0.7\linewidth}{!}{
    \begin{tabular}{c|c|c|c|c|c|c|c|c}
    \hline
    Input & Operator & ExSize & Out & SE & NL & Stride & FPN & Enhancement\\
    \hline
    $320^2\times3$ & conv2d, $3\times3$ & - & 24 & - & HS & 2 & - & c + 50\%\\
    $160^2\times24$ & bneck, $3\times3$ & 24 & 24 & - & RE & 1 & - & c + 50\%\\
    $160^2\times24$ & bneck, $3\times3$ & 72 & 36 & - & RE & 2 & - & c + 50\%\\    
    $80^2\times36$ & bneck, $3\times3$ & 108 & 36 & - & RE & 1 & - & added blocks\\
    $80^2\times36$ & bneck, $3\times3$ & 108 & 36 & - & RE & 1 & - & added blocks\\    
    $80^2\times36$ & bneck, $3\times3$ & 108 & 36 & - & RE & 1 & \checkmark & c + 50\%\\
    
    $80^2\times36$ & bneck, $5\times5$ & 108 & 60 & \checkmark & RE & 2 & - & c + 50\%\\   
    $40^2\times60$ & bneck, $5\times5$ & 180 & 60 & \checkmark & RE & 1 & - & c + 50\%\\     
    $40^2\times60$ & bneck, $5\times5$ & 180 & 60 & \checkmark & RE & 1 & \checkmark & c + 50\%\\    
    
    $40^2\times60$ & bneck, $3\times3$ & 240 & 80 & - & HS & 2 & - & -\\      
    $20^2\times80$ & bneck, $3\times3$ & 200 & 80 & - & HS & 1 & - & -\\  
    $20^2\times80$ & bneck, $3\times3$ & 184 & 80 & - & HS & 1 & - & -\\  
    $20^2\times80$ & bneck, $3\times3$ & 184 & 80 & - & HS & 1 & - & -\\  
    
    $20^2\times80$ & bneck, $3\times3$ & 480 & 112 & \checkmark & HS & 1 & - & -\\  
    $20^2\times112$ & bneck, $3\times3$ & 672 & 112 & \checkmark & HS & 1 & \checkmark & -\\  
    $20^2\times112$ & bneck, $5\times5$ & 672 & 160 & \checkmark & HS & 2 & - & -\\  
    $10^2\times160$ & bneck, $5\times5$ & 960 & 160 & \checkmark & HS & 1 & - & -\\ 
    $10^2\times160$ & bneck, $5\times5$ & 960 & 160 & \checkmark & HS & 1 & \checkmark & -\\  
    
    $10^2\times160$ & bneck, $5\times5$ & 960 & 160 & \checkmark & HS & 2 & - & -\\ 
    $5^2\times160$ & bneck, $5\times5$ & 960 & 160 & \checkmark & HS & 1 & \checkmark & -\\     
    \hline
    \end{tabular}}
\end{table*}

\section{Method}
\label{sec:framework}

In this section, we detail the design of \thename\ and illustrate how a generic detector with low FLOPs performs accurate small object detection. We follow the two-stage detection paradigm, which is more friendly to small object detection~\cite{LiuAESRFB16}. The detector contains four parts: the enhanced backbone, TinyFPN, TinyRPN and R-CNN head. For clarity, we only provide details of \thename-M in this section and leave the details of \thename-S and \thename-L in Appendix.

\subsection{Backbone Network}

\subsubsection{Detailed Information Enhancement}
\label{subsec:enhancement}
Early-stage feature maps with high resolution in the backbone contain abundant detailed information, which is vital for recognizing and localizing small objects. Existing lightweight backbone networks~\cite{HowardZCKWWAA17,SandlerHZZC18,HowardSCCCTWZPVLA19,MaZZS18,ZhangZLS18} usually downsample feature maps rapidly, keeping less layers and channels in high resolution stages. This setting successfully minimizes computational budget but sacrifices much detailed information. 

To improve the performance of detecting small objects, we propose a detailed information enhanced backbone based on the high-performance network, MobileNetV3~\cite{HowardSCCCTWZPVLA19}. Tab.~\ref{tab:backbone} shows the detailed network configuration. Compared with other widely used lightweight backbone networks~\cite{HowardZCKWWAA17,SandlerHZZC18,HowardSCCCTWZPVLA19,MaZZS18,ZhangZLS18}, 
we allocate more computation to the early stages with higher resolution. With this design, more detailed information is extracted and kept for detecting small objects.

\begin{figure}[htbp]
    \centering
    \includegraphics[width=0.98\linewidth]{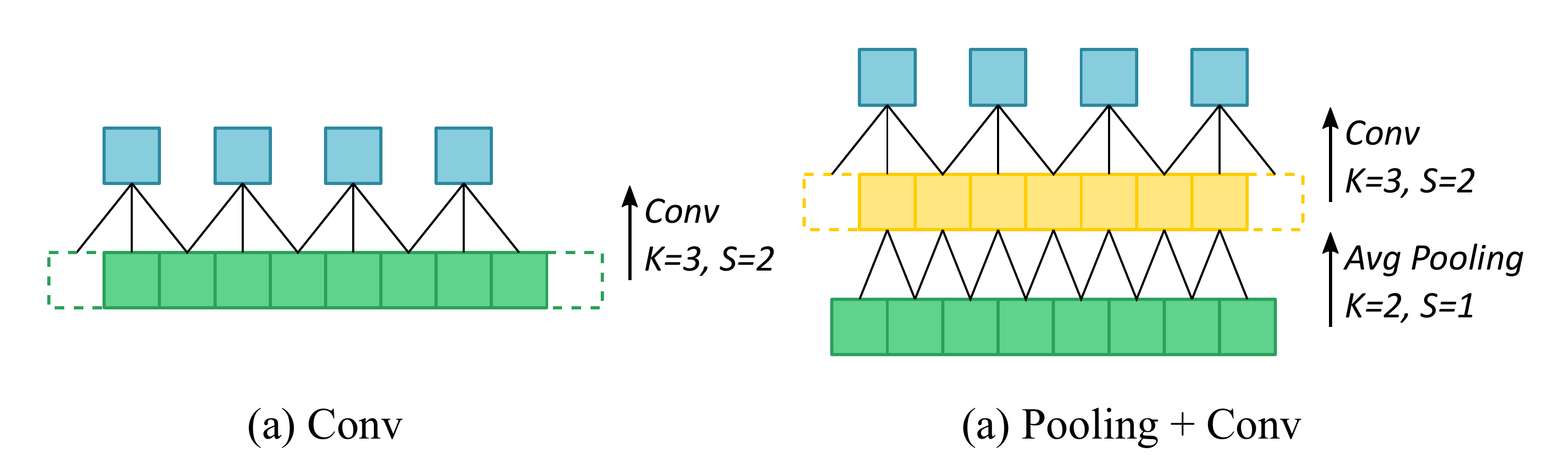}
    \caption{The feature alignment problem. (a) When the number of input pixels is even, due to spatial asymmetry, the strided convolution causes a $0.5$-pixel misalignment. (b) By introducing an average pooling before the strided convolution, even pixels are converted to odd ones, eliminating asymmetry and misalignment. Dashed squares denote zero padding.}
    \label{fig:conv_misalign_align}   
\end{figure}

\begin{figure*}[thbp]
    \centering
        \includegraphics[width=.8\linewidth]{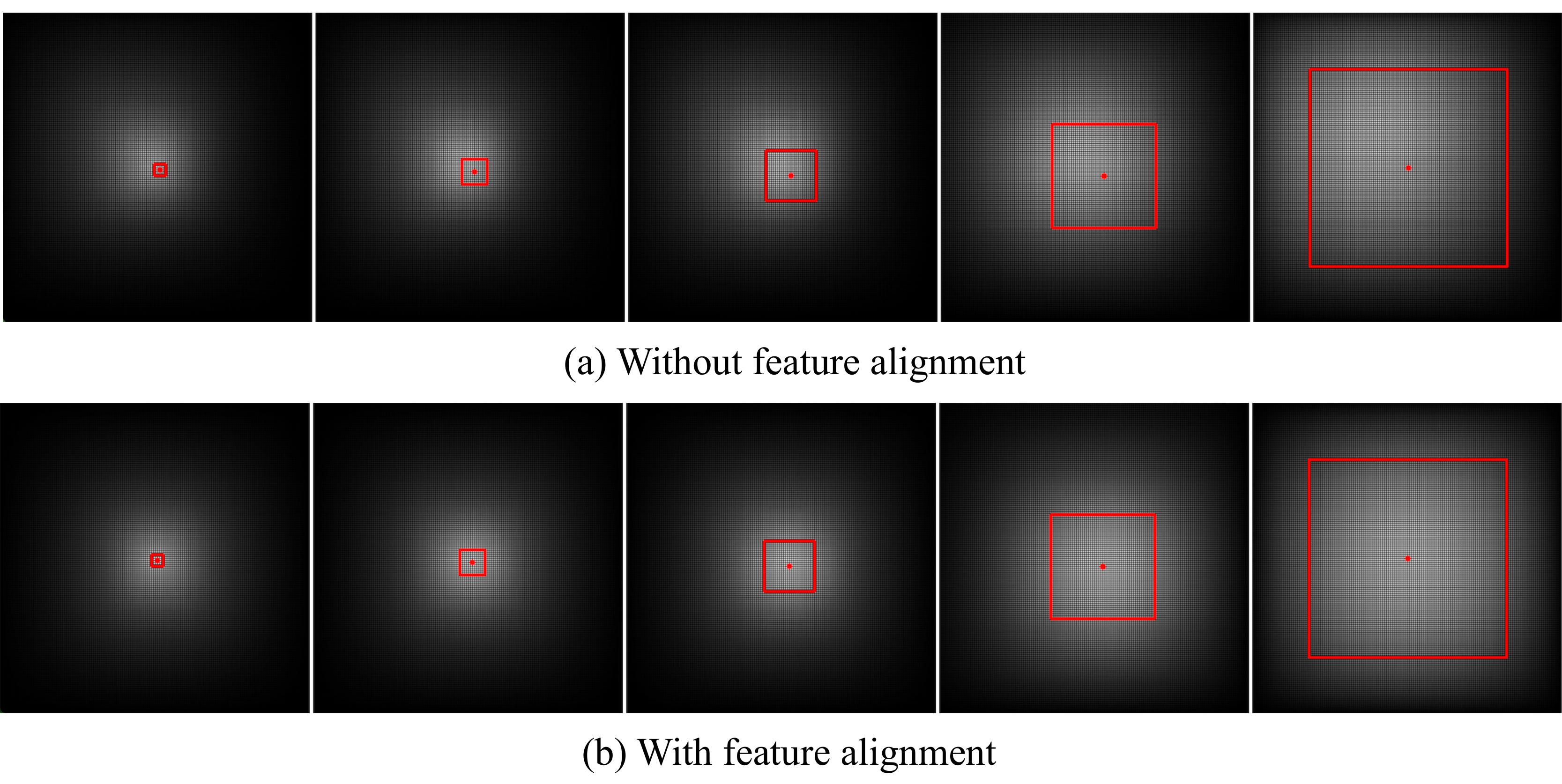}
    \caption{Visualization of ERFs of different pyramid levels. From left to right: pyramid levels with stride 4, 8, 16, 32 and 64 in FPN. The red point denotes the geometry center. The red square indicates the corresponding anchor size in each pyramid level.}
    \label{fig:ERF_misalign_align}
\end{figure*}

\subsubsection{Solving the Feature Alignment Problem}
\label{subsec:alignment}
Convolution with stride 2 is widely used to reduce the resolution of the feature maps~\cite{HowardZCKWWAA17,SandlerHZZC18,HowardSCCCTWZPVLA19,MaZZS18,ZhangZLS18}. 
And in detection, even input resolution is a common practice. Because odd input pixels ruin the proportional relation between different pyramid levels, making the coordinate mapping between input and output complex. 
However, strided convolutions on even input resolution may lead to feature misalignment.
As shown in Fig.~\ref{fig:conv_misalign_align}(a), considering a convolution layer with stride 2 and the input resolution is even, during the convolution computation, one padding pixel is ignored, resulting in spatial asymmetry and $0.5$ pixel feature misalignment. The feature misalignment caused by several strided convolutions is accumulated layer by layer through the whole network and becomes more significant in higher levels.
The misalignment is negligible for large models with high input resolution, but significant for lightweight detectors with low resolution.
It severely degrades the performance.
The RoI pooling/align operation would extract misaligned features for every object proposals. Small object detection requires more accurate localization and are affected more.

To alleviate the feature misalignment in \thename, we adopt average pooling layers before each strided convolution. As illustrated in Fig.~\ref{fig:conv_misalign_align}(b), the average pooling operation converts even pixels to odd ones and avoids asymmetry in strided convolution and correct the misalignment. And when optimizing the network, the adjacent convolution layer and averaging pooling layer can be fused into a single layer for better inference efficiency.

\paragraph{Theoretical Calculation of Feature Misalignment} We provide the theoretical calculation about the feature misalignment caused by strided convolution. For a convolution with stride $2$, we assume that the kernel size is $k$, the padding size is $\frac{k-1}{2}$, and the input feature map is with stride $s$ compared with the input image. As shown in Fig.~\ref{fig:conv_misalign_align}(a), one padding pixel is dropped, and the feature center is shifted by $0.5$ pixel. When mapped to the input image, the misalignment is $\frac{s}{2}$ pixels. The feature misalignment affects subsequent layers and is accumulated layer by layer. In \thename, $6$ strided convolution layers exist in the backbone network and respectively cause misalignment of $0.5, 1, 2, 4, 8, 16$ pixels. In FPN and RPN, the accumulated misalignment is as large as $31.5$ pixels. 
For lightweight detectors with small input size, the misalignment is quite significant. Considering the input resolution of $320\times320$, the misalignment proportion is up to $\frac{31.5}{320} \approx 9.8\%$, which leads to severe mismatching between features and their spatial positions.

\paragraph{Visualization of Effective Receptive Field} To better demonstrate the effect of feature misalignment, we adopt effective receptive field (ERF) maps~\cite{LuoLUZ16} to visualize the misalignment of feature maps. As shown in Fig.~\ref{fig:ERF_misalign_align}(a), we can observe that ERFs obviously deviate from the geometric centers of corresponding anchors. Pixel-level prediction in RPN and region-based feature extraction in R-CNN would be based on misaligned features. With average pooling layers applied, the misalignment is eliminated (Fig.~\ref{fig:ERF_misalign_align}(b)).

\begin{figure*}[!t]
\centering
\includegraphics[width=\linewidth]{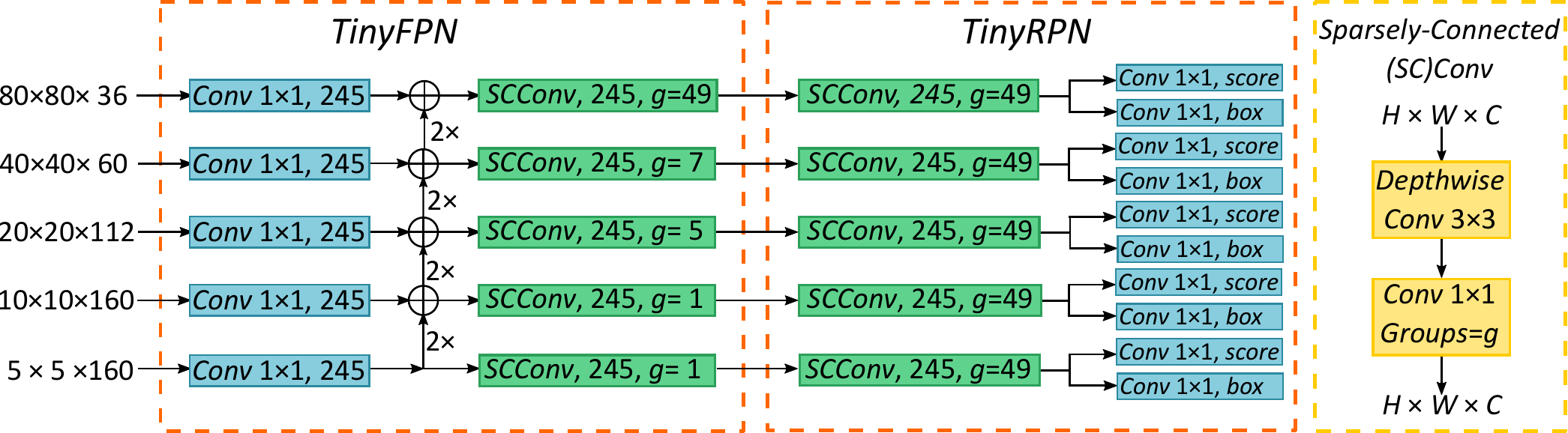}
\caption{TinyFPN and TinyRPN with sparsely-connected convolutions (SCConvs). SCConv consists of a depth-wise convolution and a point-wise group convolution. "g" denotes the number of groups in SCConv. $245$ is the number of output channels for both vanilla Conv $1\times1$ and SCConv.}
\label{fig:tiny_fpn_rpn}
\end{figure*}

\subsection{TinyFPN and TinyRPN}

\subsubsection{High Resolution Detection Feature Maps}
Under the limitation of computation cost, previous lightweight detectors usually adopt feature maps with low resolutions for detection ($38\times38$ in SSDLite~\cite{SandlerHZZC18}, $19\times19$ in Pelee~\cite{WangBL18}, $20\times20$ in ThunderNet~\cite{QinLZBYPS19}). However, small feature maps have low spatial resolution.  Low-resolution feature maps cannot provide spatially matched features for objects located in arbitrary positions, especially for small objects. In this paper, we enable object detection on high-resolution feature maps. We fetch five feature maps from the backbone for detection, respectively with stride 4, 8, 16, 32 and 64. Note that the resolution of the feature map with stride 4 is $80\times80$, which is the highest resolution feature map used in lightweight detectors. More analysis on the importance of the high-resolution setting can be found in Sec.~\ref{subsec:dense_anchor}.

\subsubsection{Sparsely-connected Convolution for Computation Reduction}
Due to the high-resolution design, computation budget of the detection part becomes extremely high. Though the depth-wise separable convolution \cite{HowardZCKWWAA17,Chollet17} has been widely used for detection part design in prior lightweight detectors~\cite{QinLZBYPS19,ChenLMXJ19,TanPL19} to reduce the computation cost, we find it not enough in our high-resolution setting. Consequently, we exploit a sparsely-connected convolution (SCConv), specialized for both efficiency and high resolution in FPN and RPN. As shown in Fig.~\ref{fig:tiny_fpn_rpn}, the SCConv is a combination of a depth-wise convolution~\cite{HowardZCKWWAA17} and a point-wise group convolution~\cite{ZhangZLS18}. Compared with the vanilla depth-wise separable convolution \cite{HowardZCKWWAA17,Chollet17}, SCConv further reduces the connections among channels. Experiments in Sec.~\ref{subsec:abalation} shows that this sparser setting has little influence on detection performance and reduces the computation cost by a large amount.

Based on SCConv, we propose TinyFPN and TinyRPN. In TinyFPN (Fig.~\ref{fig:tiny_fpn_rpn}), SCConv is applied after feature fusion, in place of the normal $3\times3$ convolutions. We set larger group numbers, \ie sparser connections, for SCConvs in the high-resolution pyramid levels to reduce the computation cost. TinyRPN (Fig.~\ref{fig:tiny_fpn_rpn}) consists of a SCConv and two sibling $1\times1$ convolutions for classification and regression respectively. The parameters of TinyRPN are shared across all pyramid levels. Ablation studies about the group settings of SCConvs are provided in Sec.~\ref{subsec:abalation}.

\begin{figure}[htb]
    \vspace{-5pt}
    \centering
    \includegraphics[width=0.98\linewidth]{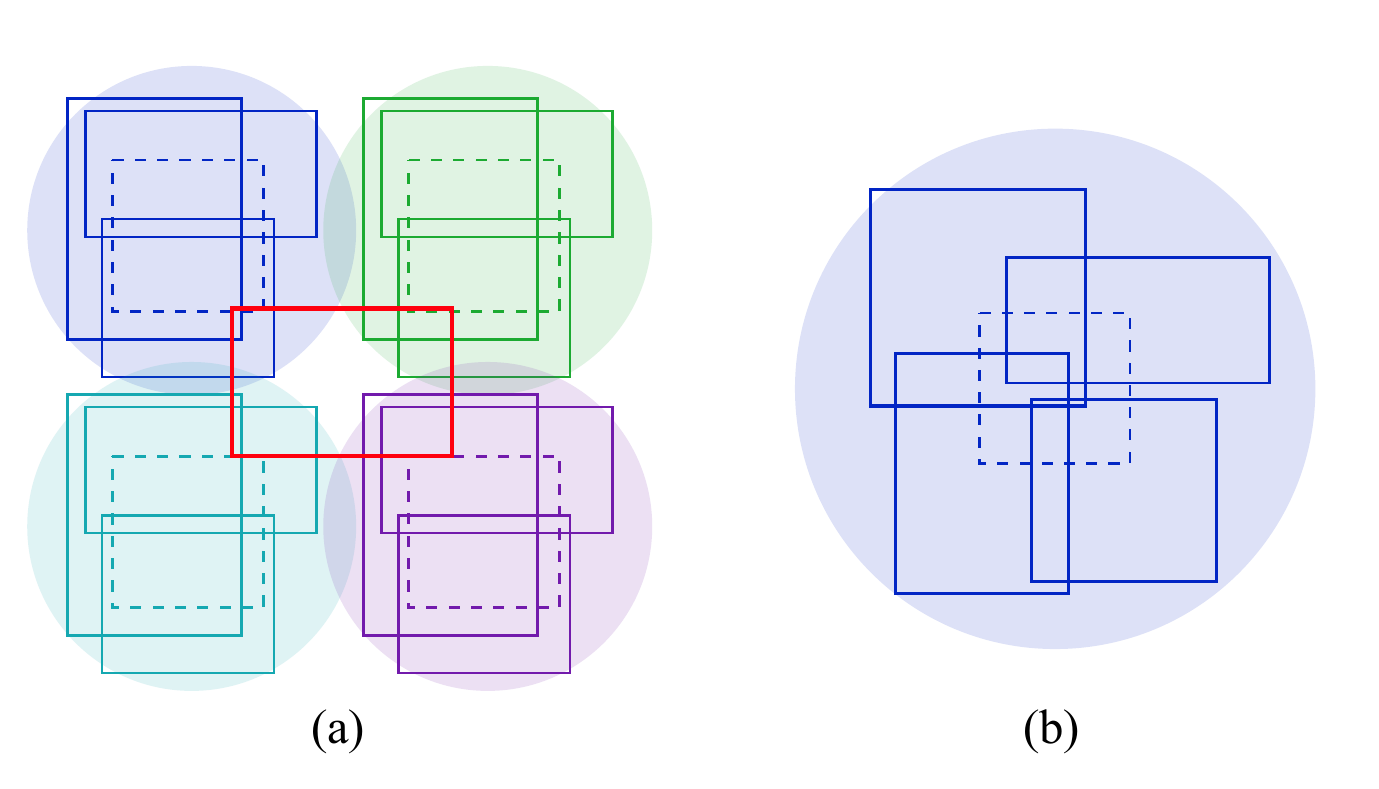}
    \caption{Problems caused by sparse anchors. Dashed squares denote anchors. Solid rectangles denote possible objects that may exist in an image. (a): Objects located in the overlooked regions (the red box) have low IoUs with every anchor, and would not be assigned to any anchor during training. (b): With the responsive region enlarged, every anchor deals with more objects and more variance in object shape and position. Detection becomes harder.}
    \label{fig:anchor_problem}
\end{figure}

\begin{figure}[htb]
\centering
\includegraphics[width=0.98\linewidth]{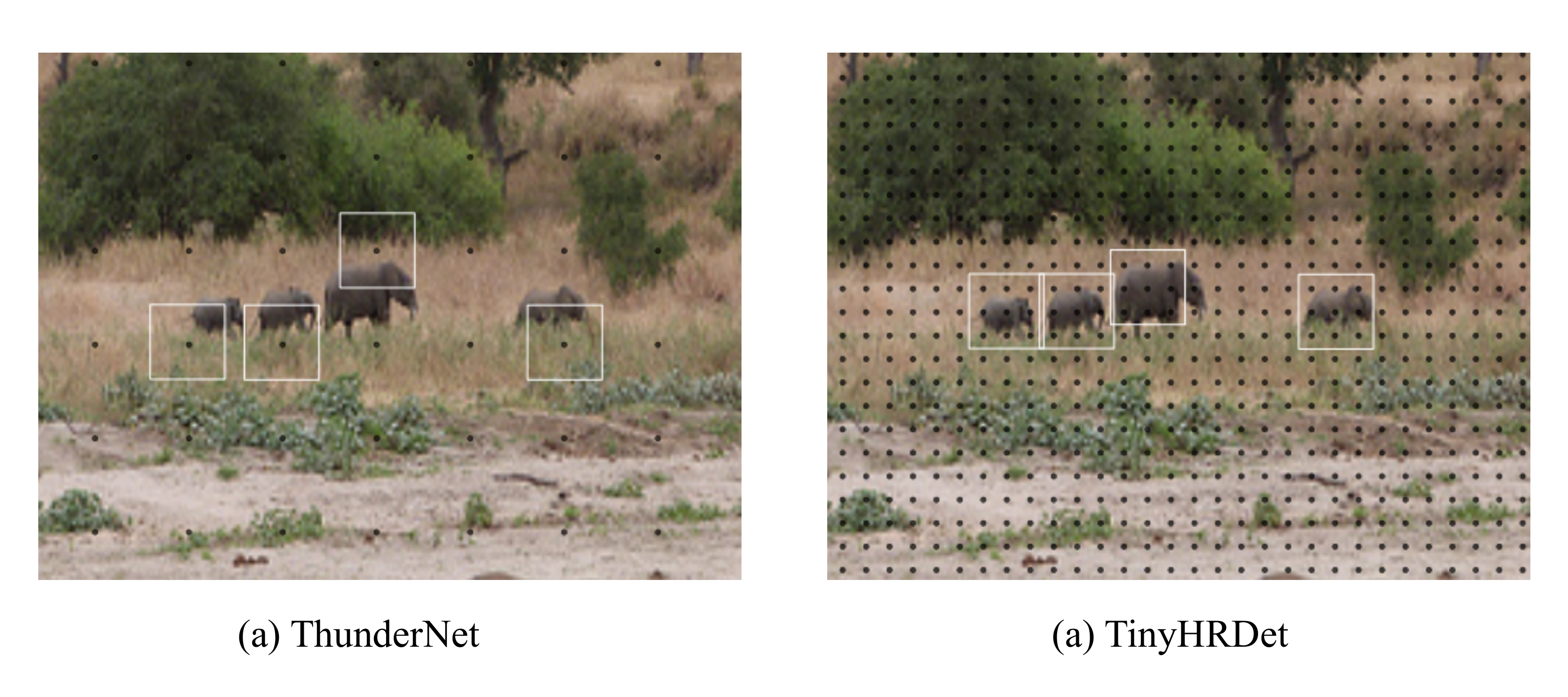}
\caption{Anchor density comparison between ThunderNet and \thename. Black points denote centers of evenly tiled anchors and white boxes show the anchors that have the best IoU with a ground-truth object. Densely tiled anchors better cover small objects.}
\label{fig:density_comp}
\end{figure}

\subsection{High-resolution Feature Maps for Dense Anchoring}
\label{subsec:dense_anchor}
Translation variance is a challenge in object detection. Detectors should be insensitive to translation and deal with objects located in arbitrary positions. The introduction of anchor~\cite{RenHGS15} eases the problem of translation variance in object detection. A large number of anchors are evenly tiled over the whole image and each anchor is only responsible for predicting objects that appear in a certain region, which is defined as \emph{responsive region}. During training, objects are assigned to anchors according to their IoUs and an assignment strategy~\cite{RenHGS15}. However, two problems emerge in the procedure of anchor assignment when anchors are not spatially dense enough, which are described as follows.
\begin{itemize}
    \item With the assignment IoU threshold fixed, responsive regions are fixed. As is shown in Fig.~\ref{fig:anchor_problem}(a), when anchors are not dense enough, responsive regions cannot cover the whole image. We call the uncovered region \emph{overlooked region}. Objects, especially small objects, located in overlooked regions are never assigned to any anchor during training. Consequently, in the inference phase, these objects are less possible to be detected for lack of anchors here. 
    \item If we lower the assignment IoU threshold, the responsive regions are enlarged while the overlooked regions shrink or even vanish. But enlarged responsive regions make it hard for detectors to obtain accurate results. Each anchor deals with more objects and more variance in object shape and position, which is illustrated in Fig.~\ref{fig:anchor_problem}(b).
\end{itemize}

As discussed above, the anchor density should be high enough to cover possible objects.
Thus, in \thename, we keep high-resolution feature maps for dense  anchoring. 
To comprehend how dense anchoring impact detection performance, especially for small objects, we take  \thename\ with ThunderNet~\cite{QinLZBYPS19} for comparison. In Fig.~\ref{fig:density_comp}, we visualize the distribution of smallest anchors. In ThunderNet, the distance between adjacent anchors is $16$ pixels. It's hard to match small objects with anchors spatially. While that in \thename\ is only $4$ pixels. Densely tiled anchors better cover small objects.

\begin{table}[htb]
    \centering
    \caption{Ground-truth miss-assignment ratio (GTMR) for Faster R-CNN, ThunderNet and our \thename. GTMR$^s$, GTMR$^m$ and GTMR$^l$ for small, medium and large objects respectively.}
    \label{tab:miss_ratio}
    \resizebox{0.98\linewidth}{!}{
    \begin{tabular}{l c c c c c}
    \toprule
    Detector & GTMR(\%) & GTMR$^s$(\%) & GTMR$^m$(\%) & GTMR$^l$(\%) & Input\\
    \midrule
    Faster R-CNN  & $6.6$ & 17.0 & 1.2 & 1.9 & $800\times1333$\\
    Faster R-CNN w/ FPN & $6.1$ &  15.2 & 1.2 & 2.3 & $800\times1333$\\
    ThunderNet & $18.1$ & 53.1 & 5.6& 1.2 & $320\times320$\\
    \thename & $8.4$ & 21.2 & 2.2 & 3.0 &$320\times320$\\
    \bottomrule
    \end{tabular}
    }
\end{table}

Quantitatively, we propose \emph{ground-truth miss-assignment ratio} (GTMR) to evaluate the assignment procedure. GTMR is defined as the proportion of those ground-truth objects that are not assigned to any anchor over all the objects. It reflects the matching quality between anchors and objects under a certain assignment strategy. As shown in Tab.~\ref{tab:miss_ratio}, GTMR of ThunderNet is up to $18.1\%$, much higher than that of Faster R-CNN~\cite{RenHGS15}. \thename\ obtains a quite lower GTMR, $8.4\%$, though as lightweight as ThunderNet. Notably, GTMR of small objects of ThunderNet is extremely high; while, in our \thename\ with high-resolution feature maps and densely tiled anchors, GTMR of small objects is much lower than ThunderNet. Small objects can be better matched with anchors spatially.

\section{Experiments}
In this section, we first describe the experimental details. Then we compare our \thename\ models with other state-of-the-art (SOTA) methods. We further perform detailed ablation studies to demonstrate the effectiveness of our proposed methods.

\subsection{Experimental Details} 
Our experiments are conducted on the COCO dataset \cite{LinMBHPRDZ14}. We use the \textit{train2017} split for training and report our main results on the \textit{test-dev2017}. The \textit{val2017} split is used for detailed ablation studies. We follow the standard COCO detection metrics to report the average precision (AP) under different IoUs (\ie, AP$^{50}$, AP$^{75}$ and the overall AP) and AP for detecting objects in different scales (\ie, AP$^s$, AP$^m$ and AP$^l$).

\begin{table*}[!t]
    \caption{SOTA results on the COCO \textit{test-dev2017} set. All the results are obtained via the single-model and single-scale testing. $^\dag$ means the model is re-implemented by ourselves. $^\ddag$ denotes the result is evaluated on \textit{val2017} set because the result on \textit{test-dev2017} set is not provided.}
    \centering
    \scriptsize
    \resizebox{0.9\linewidth}{!}{
    \begin{tabular}{l|c|c c c|c c c|c}
    \toprule
    Method & FLOPs & AP  & AP$^{50}$ & AP$^{75}$ & AP$^{s}$ & AP$^{m}$ & AP$^{l}$ & Input\\
    \midrule
    ThunderNet-SNet146~\cite{QinLZBYPS19} & 470M & 23.6 & 40.2 & 24.5 & - & - & - &$320^2$\\
    ThunderNet-SNet146$^\dag$ & 499M & 23.8 & 40.5 & 24.7 & 4.6 & 23.0 & 42.9 & $320^2$\\
    \thename-S & 495M & 26.0 & 45.8 & 26.5 & 9.6 &  26.8 & 39.5 & $320^2$\\
    \midrule
    MobileNetV2-SSDLite~\cite{SandlerHZZC18} & 800M & 22.1 & - & - & - & - & - &$320^2$\\
    MobileNet-SSDLite~\cite{SandlerHZZC18} & 1300M & 22.2 & - & - & - & - & - &$320^2$\\
    Pelee~\cite{WangBL18} & 1290M & 22.4 & 38.3 & 22.9 & - & - & - &$304^2$\\
    Tiny-DSOD~\cite{LiLLL18} & 1120M & 23.2 & 40.4 & 22.8 & - & - & - &$300^2$\\
    YOLOX-Nano$^\ddag$~\cite{YOLOX} & 1080M & 25.3 & - & - & - & - & - & $416^2$\\
    NASFPNLite MobileNetV2~\cite{GhiasiLL19} & 980M & 25.7& - & - & - & - & - & $320^2$\\
    ThunderNet-SNet535~\cite{QinLZBYPS19} & 1300M & 28.0  & 46.2 & 29.5 & - & - & - &$320^2$\\
    ThunderNet-SNet535$^\dag$ & 1297M & 27.3 & 45.4 & 28.4 & 6.5 & 28.4 & 46.2  & $320^2$\\
    \thename-M & 991M & 30.3 & 51.2 & 31.8 & 13.5 & 30.9 & 43.9 & $320^2$\\
    
    \midrule

    YOLOv2~\cite{RedmonF17} & 17.5G & 21.6 & 44.0 & 19.2 & 5.0 & 22.4 & 35.5 & $416^2$\\
    PRN~\cite{PRN} & 4.0G &23.3 & 45.0 & 22.0 & 6.7 & 24.8 & 35.1 &$416^2$\\
    EFM (SAM)~\cite{EFM} & 5.1G & 26.8 & 49.0 & 26.7 & 9.8 & 28.2 & 38.8 & $416^2$\\
    YOLOX-Tiny$^\ddag$~\cite{YOLOX} & 6.5G & 31.7 & - & - & - & - & - & $416^2$\\
    EfficientDet-D0~\cite{TanPL19} & 2.5G & 32.4 & - & - & - & - & - &$512^2$\\
    YOLOv3~\cite{RedmonF18} & 71.0G & 33.0 & 57.9 & 34.4 & 18.3 & 35.4 & 41.9 &$608^2$\\
    YOLOv5~\cite{YOLOv5} & 17.0G & 36.7 & - & - & - & - & - & $640^2$\\
    \thename-L & 2.4G & 35.5 & 56.8 & 38.3 & 18.3 & 37.5 & 48.4 &$512^2$\\
    \bottomrule
    \end{tabular}
    }

    \label{tab:SOTA}
\end{table*}

We implement our \thename\ based on the PyTorch~\cite{paszke2019pytorch} framework and MMDetection~\cite{mmdetection} toolbox.  Our models are trained with batch size $128$ on $4$ GPUs ($32$ images per GPU) for $240$ epochs. The SGD optimizer is used with momentum $0.9$ and weight decay 1e-5. We linearly increase the learning rate from $0$ to $0.35$ in the first $500$ iterations and then decay it to 1e-5 using the cosine anneal schedule~\cite{loshchilov2016sgdr}. $2000$/$200$ proposals are used in the second stage at the training/inference phase. We adopt the same data augmentation strategy introduced in SSD~\cite{LiuAESRFB16}. Following ThunderNet~\cite{QinLZBYPS19}, online hard example mining (OHEM)~\cite{ShrivastavaGG16}, Soft-NMS~\cite{BodlaSCD17}, and Cross-GPU Batch Normalization~\cite{PengXLJZJYS18} are used. 

In the proposed TinyRPN head, anchors in five pyramid levels are respectively with sizes of $12.8^2$, $25.6^2$, $51.2^2$, $102.4^2$ and $204.8^2$. Anchors of multiple aspect ratios $\{1:2,\ 1:1,\ 2:1\}$ are used in each level, the same as \cite{RenHGS15} and \cite{LinDGHHB17}. As for the R-CNN head, we adopt position-sensitive RoI align~\cite{DaiLHS16,LiPYZDS17,QinLZBYPS19} to extract box features effectively.

\subsection{Comparison with SOTA Lightweight Generic Detectors}
We compare our \thename\ with other SOTA lightweight detectors on the COCO \textit{test-dev2017} set in Tab.~\ref{tab:SOTA}. Among the compared methods, ThunderNet~\cite{QinLZBYPS19} and EfficientDet~\cite{TanPL19} are regarded as the most recent SOTA lightweight detectors. The results show that our models obviously outperform them with fewer computation costs: \thename-S surpasses ThunderNet-SNet146 by $2.4\%$ AP with similar FLOPs; \thename-M surpasses ThunderNet-SNet535 by $2.3\%$ AP with $76\%$ FLOPs; and \thename-L surpasses EfficientDet-D0 by $3.1\%$ AP with similar FLOPs. Besides, \thename\ has better performance-computation trade-offs than the automatically searched lightweight detector (\ie, NAS-FPNLite~\cite{GhiasiLL19}) and the popular YOLO series~\cite{RedmonF17,RedmonF18,YOLOv5}.

\begin{figure*}[h]
    \centering
    \includegraphics[width=1\linewidth]{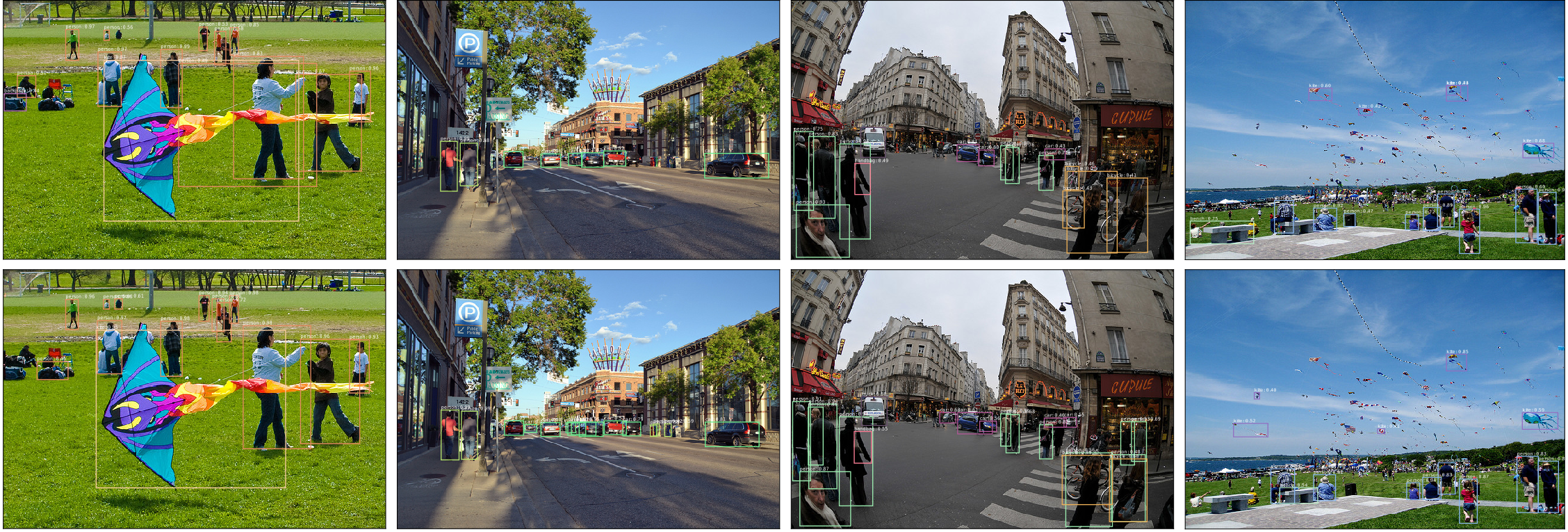}
    \caption{Some visualized detection results of \thename-S in the top row and \thename-M in the bottom row. Best view in PDF.}
        \label{fig:somevis}   
\end{figure*}

Our \thename\ models obtain extraordinary performance on detecting small objects. With similar computation cost (\ie, FLOPs), \thename-M achieves $13.5$ AP$^{s}$, which is over $100\%$ improvement over ThunderNet-SNet535 with $6.5$ AP$^{s}$; and \thename-S achieves $9.6$ AP$^{s}$, which is also over $100\%$ improvement over ThunderNet-SNet146 with $4.6$ AP$^{s}$. \thename-L achieves the same $18.3$ AP$^s$ with YOLOv3~\cite{RedmonF18}, but it only has $1/30$ computation cost of YOLOv3. Some visualized results are shown in Fig.~\ref{fig:somevis}.

\subsection{Ablation Studies}
\label{subsec:abalation}

\begin{table}[t!]
    \caption{Evaluation of backbone enhancement. The enhanced backbone significantly improves the detection performance, especially that of small objects. }
    \centering
    \resizebox{0.98\linewidth}{!}{
    \begin{tabular}{l|c|c c c c}
    \toprule
    Backbone & MFLOPs & AP  & AP$^{s}$ & AP$^{m}$ & AP$^{l}$\\
    \midrule
    MobileNetV2~\cite{SandlerHZZC18} & 908 & 28.2 & 12.5 & 29.5 & 43.3\\    
    SNet535~\cite{QinLZBYPS19} & 1602 & 28.5 & 11.9 & 29.3 & 44.9\\
    FBNet-C~\cite{WuDZWSWTVJK19} & 1055 & 27.8 & 11.6 & 28.6 & 43.9\\
    Proxyless-GPU~\cite{CaiZH19} & 1304 & 27.8 & 11.8 & 28.4 & 44.3\\
    MobileNetV3-BC w/o SE & 988 & 29.5 & 14.2 & 31.1 & 44.1 \\
    \midrule
    MobileNetV3~\cite{HowardSCCCTWZPVLA19} & 963  & 28.4 & 13.0 & 29.6 & 43.7 \\
    MobileNetV3-B  & 1016 & 29.5 & 14.3 & 30.7 & 44.3 \\
    MobileNetV3-C  & 1001 & 29.8 & 14.6 & 30.9 & 44.8 \\
    MobileNetV3-BC & 991  & 30.3 & 15.6 & 31.9 & 45.0 \\
    \bottomrule
    \end{tabular}}

    \label{tab:backbone_abalation}
\end{table}

\subsubsection{Ablation Study on Backbone Enhancement}
As discussed in Sec.~\ref{subsec:enhancement}, we enhance detailed information in the MobileNetV3-based backbone, which is important for improving the feature representation of high resolution. We provide three MobileNetV3's variants: MobileNetV3-B, -C and -BC to demonstrate our effectiveness of the enhanced backbone. Compared with original MobileNetV3, MobileNetV3-B adds $2$ extra blocks in the stage with stride 4 and MobileNetV3-C contains $50\%$ more channels in early stages. MobileNet-BC is the combination of MobileNetV3-B and MobileNetV3-C which is our proposed backbone for \thename\ (configurations in Tab.~\ref{tab:backbone}). Tab.~\ref{tab:backbone_abalation} shows the comparison of different configurations. For the fair comparison, we adjust group numbers of TinyFPN and TinyRPN to keep the total computation budget similar (around $1$ GFLOPs). Our MobileNetV3-BC outperforms MobileNetV3 and other variants. Increasing layer numbers and channel numbers respectively improves AP by $1.1\%$ and $1.4\%$. When both are adopted, AP is improved  by $1.9\%$. Besides, the gain on AP$^{s} (+2.6)$ is more significant than AP$^{m} (+1.3)$ and AP$^{l} (+1.3)$. With more computation allocated to the early stages, detailed information is enhanced and benefits small object detection more.

We also compare MobileNetV3-BC with other representative lightweight backbones (Tab.~\ref{tab:backbone_abalation}). We remove SE module for fair comparison. MobileNetV3-BC achieves better results than both manually designed backbones (SNet535~\cite{QinLZBYPS19} and MobileNetV2~\cite{SandlerHZZC18}) and automatically searched backbones (FBNet~\cite{WuDZWSWTVJK19} and Proxyless~\cite{CaiZH19}).

\subsubsection{Ablation Study on the SCConv in TinyFPN and TinyRPN}
To make FPN and RPN lightweight, we combine the depth-wise convolution and point-wise group convolution as the sparsely-connected convolution (SCConv). As shown in Tab.~\ref{tab:sparse_cfg}, compared with the conventional convolution or depth-wise separable convolution~\cite{HowardZCKWWAA17,Chollet17} in FPN and RPN, SCConv greatly reduces the computation cost and achieves comparable results. The results confirm that the SCConv in TinyFPN and TinyRPN causes little degradation on detection performance. For the lightweight detector design, adopting SCConv in detection heads is an effective method to save computation cost while maintaining high detection performance.

\begin{table}[!t]
    \caption{Results of applying different convolution operators in FPN and RPN. The proposed sparsely-connected convolution in FPN and RPN greatly reduces the computation cost and keeps comparable performance.}
    \centering
    \resizebox{0.98\linewidth}{!}{
    \begin{tabular}{l  c  c c c}
    \toprule
    Conv type & MFLOPs & AP  & AP$^{50}$ & AP$^{75}$\\
    \midrule
    Conventional Conv & 10115 & 31.8 & 52.5 & 33.5 \\
    Depth-wise Separable Conv & 1970 & 30.8 & 51.5 & 32.4 \\
    Sparsely-connected Conv & 991 & 30.3 & 50.9 & 32.1 \\
    \bottomrule
    \end{tabular}}

    \label{tab:sparse_cfg}
\end{table}

We also evaluate different group number settings of SCConvs in TinyFPN and TinyRPN. The results are presented in Tab.~\ref{tab:group_cfg}. From the comparison between row (c) and (e), we find that setting different group numbers for different pyramid levels is better than setting the same group number. A good practice is to use more groups for the bottom pyramids and fewer groups in the top pyramids in FPN (Tab.~\ref{tab:group_cfg}(e)). 
 
\begin{table}[htbp]
    \caption{Evaluation of different group configurations. TinyFPN groups in the table respectively corresponds to pyramid levels with stride 4, 8, 16, 32 and 64. We adopt (e) in our \thename.}
    \centering
    \resizebox{0.98\linewidth}{!}{
    \setlength{\tabcolsep}{1pt}{
    \begin{tabular}{c c c c c c c}
    \toprule
    & TinyFPN groups & TinyRPN group & MFLOPs & AP  & AP$^{50}$ & AP$^{75}$\\
    \midrule
    (a) &1, 1, 1, 1, 1 & 1 & 1970 & 30.8 & 51.5 & 32.4 \\
    (b) &245, 245, 245, 245, 245 & 245 & 950 & 28.8 & 49.6 & 30.1\\
    (c) &7, 7, 7, 7, 7 & 49 & 1030 & 30.0 & 50.4 & 31.4 \\
    (d) &1, 1, 5, 7, 49 & 49 & 1440 & 30.6 & 51.1 & 32.4\\
    (e) &49, 7, 5, 1, 1 & 49 & 991 & 30.3 & 50.9 & 32.1 \\
    \bottomrule
    \end{tabular}}
    }
    \label{tab:group_cfg}
\end{table}

\subsubsection{Ablation Study on Feature Alignment}
We evaluate how feature misalignment, described in Sec.~\ref{subsec:alignment}, affects the detection performance in Tab.~\ref{tab:misalignment}. By better aligning the backbone features using a simple average pooling, we get a $0.5\%$ gain on AP. Note that the gain on AP$^{s}$ ($+1.4\%$) is higher than that on AP$^{m}$ ($+0.6\%$) or AP$^{l}$ ($+0.8\%$). This is because small objects are much more sensitive to feature misalignment and benefit more from this correction.

\begin{table}[thbp]
    \caption{Detection results without and with feature alignment. Alignment significantly improves detection performance of small objects.}
    \centering
    \setlength{\tabcolsep}{1pt}{
    \resizebox{0.98\linewidth}{!}{
    \begin{tabular}{l l l l l l l}
    \toprule
     & AP  & AP$^{50}$ & AP$^{75}$ & AP$^{s}$ & AP$^{m}$ & AP$^{l}$ \\
    \midrule
    without alignment & 29.8 & 50.4 & 31.1 & 14.2 & 31.3 & 44.2\\
    with alignment & 30.3$_{\uparrow0.5}$ & 50.9$_{\uparrow0.5}$ & 32.1$_{\uparrow1.0}$ & 15.6$_{\uparrow1.4}$ & 31.9$_{\uparrow0.6}$ & 45.0$_{\uparrow0.8}$\\
    \bottomrule
    \end{tabular}}
    }
    \label{tab:misalignment}
\end{table}

\begin{table}[h]
\centering
\caption{Inference latency of TinyDet on two types of ARM CPUs with single thread. No specific optimization is adopted.}
\label{tab:latency}
\resizebox{0.7\linewidth}{!}{
\begin{tabular}{lcc}
\toprule
CPU type & Snapdragon 865 & Kirin 820\\
\midrule 
TinyDet-S & 103ms & 133ms\\
TinyDet-M & 179ms & 236ms\\
TinyDet-L & 312ms & 386ms\\
\bottomrule
\end{tabular}
}
\end{table}

\subsection{Limitation} There is a limitation in this study. 
Currently we cannot fairly compare the speed with other detectors.
The reasons are two-fold.
Firstly, the speed of a detector is highly dependent on the inference SDK (\eg, Apple's CoreML and Google's TensorFlowLite). For ThunderNet, it uses a third-party high-performance inference SDK, which is not publicly available. This makes fair speed comparison in the same hardware condition infeasible.
Secondly, most inference SDKs are mainly optimized for computation-intensive models and less focuses on the memory bandwidth. However, \thename\ is computation-efficient and requires bandwidth-oriented optimization. The inconsistency makes the \thename\ sub-optimal in speed. Currently we offer the inference latency of TinyDet on two types of ARM CPUs with single thread and without any optimization (Tab.~\ref{tab:latency}).

FLOPs is a widely-accepted metric of great generality, especially in the neural architecture search (NAS) domain. FLOPs can be calculated theoretically, irrelevant to specific hardware implementation. And according to the roofline theory~\cite{Roofline}, FLOPs determines the upper bound of speed.
Considering the low computation requirement, with specific optimization, \thename\ can achieve excellent inference speed.

\section{Conclusion}
In this paper, we target at detecting generic (\ie, 80 object categories) small (\ie, $\leq (32\times32)/(480\times640) \approx 0.33\%$ area of the input image)\footnote{According to the definition of small objects in the COCO dataset, referring to \cite{LinMBHPRDZ14} and \url{http://cocodataset.org/\#detection-eval}} objects in the computation constrained (\ie, $\leq$ 1 GFLOPs) setting. The proposed technologies, such as high-resolution detection feature maps, TinyFPN/RPN, early-stage enhanced backbone and average pooling-based feature alignment are focusing on solving this problem. We obtain outstanding results - more than 100\% AP improvement over the previous state-of-the-art lightweight generic detector on COCO small object detection. In the future, we would like to implement and optimize the proposed \thename\ on ASICs and edge devices.



{\small
\bibliographystyle{ieee_fullname}
\bibliography{egbib}

\begin{thebibliography}{10}\itemsep=-1pt

\bibitem{BaiZDG18}
Yancheng Bai, Yongqiang Zhang, Mingli Ding, and Bernard Ghanem.
\newblock {SOD-MTGAN:} small object detection via multi-task generative
  adversarial network.
\newblock In {\em {ECCV}}, 2018.

\bibitem{BellZBG16}
Sean Bell, C.~Lawrence Zitnick, Kavita Bala, and Ross~B. Girshick.
\newblock Inside-outside net: Detecting objects in context with skip pooling
  and recurrent neural networks.
\newblock In {\em {CVPR}}, 2016.

\bibitem{BodlaSCD17}
Navaneeth Bodla, Bharat Singh, Rama Chellappa, and Larry~S. Davis.
\newblock Soft-nms - improving object detection with one line of code.
\newblock In {\em {ICCV}}, 2017.

\bibitem{CaiZH19}
Han Cai, Ligeng Zhu, and Song Han.
\newblock Proxylessnas: Direct neural architecture search on target task and
  hardware.
\newblock In {\em {ICLR}}, 2019.

\bibitem{CaoXYLSW2017}
Guimei Cao, Xuemei Xie, Wenzhe Yang, Quan Liao, Guangming Shi, and Jinjian Wu.
\newblock Feature-fused {SSD:} fast detection for small objects.
\newblock {\em arXiv:1709.05054}, 2017.

\bibitem{ChenLMXJ19}
Chen Chen, Mengyuan Liu, Xiandong Meng, Wanpeng Xiao, and Qi Ju.
\newblock Refinedetlite: {A} lightweight one-stage object detection framework
  for cpu-only devices.
\newblock {\em arXiv:1911.08855}, 2019.

\bibitem{mmdetection}
Kai Chen, Jiaqi Wang, Jiangmiao Pang, Yuhang Cao, Yu Xiong, Xiaoxiao Li,
  Shuyang Sun, Wansen Feng, Ziwei Liu, Jiarui Xu, Zheng Zhang, Dazhi Cheng,
  Chenchen Zhu, Tianheng Cheng, Qijie Zhao, Buyu Li, Xin Lu, Rui Zhu, Yue Wu,
  Jifeng Dai, Jingdong Wang, Jianping Shi, Wanli Ouyang, Chen~Change Loy, and
  Dahua Lin.
\newblock {MMDetection}: Open mmlab detection toolbox and benchmark.
\newblock {\em arXiv:1906.07155}, 2019.

\bibitem{Chollet17}
Fran{\c{c}}ois Chollet.
\newblock Xception: Deep learning with depthwise separable convolutions.
\newblock In {\em {CVPR}}, 2017.

\bibitem{DaiLHS16}
Jifeng Dai, Yi Li, Kaiming He, and Jian Sun.
\newblock {R-FCN:} object detection via region-based fully convolutional
  networks.
\newblock In {\em {NIPS}}, 2016.

\bibitem{FuLRTB17}
Cheng{-}Yang Fu, Wei Liu, Ananth Ranga, Ambrish Tyagi, and Alexander~C. Berg.
\newblock {DSSD} : Deconvolutional single shot detector.
\newblock {\em arXiv:1701.06659}, 2017.

\bibitem{YOLOX}
Zheng Ge, Songtao Liu, Feng Wang, Zeming Li, and Jian Sun.
\newblock Yolox: Exceeding yolo series in 2021.
\newblock {\em arXiv preprint arXiv:2107.08430}, 2021.

\bibitem{GhiasiLL19}
Golnaz Ghiasi, Tsung{-}Yi Lin, and Quoc~V. Le.
\newblock {NAS-FPN:} learning scalable feature pyramid architecture for object
  detection.
\newblock In {\em {CVPR}}, 2019.

\bibitem{GoodfellowPMXWOCB14}
Ian~J. Goodfellow, Jean Pouget{-}Abadie, Mehdi Mirza, Bing Xu, David
  Warde{-}Farley, Sherjil Ozair, Aaron~C. Courville, and Yoshua Bengio.
\newblock Generative adversarial nets.
\newblock In {\em NIPS}, 2014.

\bibitem{HowardSCCCTWZPVLA19}
Andrew Howard, Mark Sandler, Grace Chu, Liang{-}Chieh Chen, Bo Chen, Mingxing
  Tan, Weijun Wang, Yukun Zhu, Ruoming Pang, Vijay Vasudevan, Quoc~V. Le, and
  Hartwig Adam.
\newblock Searching for mobilenetv3.
\newblock {\em arXiv:1905.02244}, 2019.

\bibitem{HowardZCKWWAA17}
Andrew~G. Howard, Menglong Zhu, Bo Chen, Dmitry Kalenichenko, Weijun Wang,
  Tobias Weyand, Marco Andreetto, and Hartwig Adam.
\newblock Mobilenets: Efficient convolutional neural networks for mobile vision
  applications.
\newblock {\em arXiv:1704.04861}, 2017.

\bibitem{hu2018squeeze}
Jie Hu, Li Shen, and Gang Sun.
\newblock Squeeze-and-excitation networks.
\newblock In {\em CVPR}, 2018.

\bibitem{HuR17}
Peiyun Hu and Deva Ramanan.
\newblock Finding tiny faces.
\newblock In {\em {CVPR}}, 2017.

\bibitem{YOLOv5}
Glenn Jocher, Yonghye Kwon, guigarfr, perry0418, Josh Veitch-Michaelis, Ttayu,
  Daniel Suess, Fatih Baltacı, Gabriel Bianconi, IlyaOvodov, Marc, e96031413,
  Chang Lee, Dustin Kendall, Falak, Francisco Reveriano, FuLin, GoogleWiki,
  Jason Nataprawira, Jeremy Hu, LinCoce, LukeAI, NanoCode012, NirZarrabi,
  Oulbacha Reda, Piotr Skalski, SergioSanchezMontesUAM, Shiwei Song, Thomas
  Havlik, and Timothy~M. Shead.
\newblock {ultralytics/yolov3: v9.5.0 - YOLOv5 v5.0 release compatibility
  update for YOLOv3}, Apr. 2021.

\bibitem{LiLWXFY17}
Jianan Li, Xiaodan Liang, Yunchao Wei, Tingfa Xu, Jiashi Feng, and Shuicheng
  Yan.
\newblock Perceptual generative adversarial networks for small object
  detection.
\newblock In {\em {CVPR}}, 2017.

\bibitem{LiLLL18}
Yuxi Li, Jiuwei Li, Weiyao Lin, and Jianguo Li.
\newblock Tiny-dsod: Lightweight object detection for resource-restricted
  usages.
\newblock In {\em {BMVC}}, 2018.

\bibitem{LiPYZDS17}
Zeming Li, Chao Peng, Gang Yu, Xiangyu Zhang, Yangdong Deng, and Jian Sun.
\newblock Light-head {R-CNN:} in defense of two-stage object detector.
\newblock {\em arXiv:1711.07264}, 2017.

\bibitem{Li_2018_ECCV}
Zeming Li, Chao Peng, Gang Yu, Xiangyu Zhang, Yangdong Deng, and Jian Sun.
\newblock Detnet: Design backbone for object detection.
\newblock In {\em The European Conference on Computer Vision (ECCV)}, September
  2018.

\bibitem{LinDGHHB17}
Tsung{-}Yi Lin, Piotr Doll{\'{a}}r, Ross~B. Girshick, Kaiming He, Bharath
  Hariharan, and Serge~J. Belongie.
\newblock Feature pyramid networks for object detection.
\newblock In {\em {CVPR}}, 2017.

\bibitem{LinGGHD17}
Tsung{-}Yi Lin, Priya Goyal, Ross~B. Girshick, Kaiming He, and Piotr
  Doll{\'{a}}r.
\newblock Focal loss for dense object detection.
\newblock In {\em {ICCV}}, 2017.

\bibitem{LinMBHPRDZ14}
Tsung{-}Yi Lin, Michael Maire, Serge~J. Belongie, James Hays, Pietro Perona,
  Deva Ramanan, Piotr Doll{\'{a}}r, and C.~Lawrence Zitnick.
\newblock Microsoft {COCO:} common objects in context.
\newblock In {\em {ECCV}}, 2014.

\bibitem{LiuWFTP15}
Baoyuan Liu, Min Wang, Hassan Foroosh, Marshall~F. Tappen, and Marianna Pensky.
\newblock Sparse convolutional neural networks.
\newblock In {\em {CVPR}}, 2015.

\bibitem{LiuAESRFB16}
Wei Liu, Dragomir Anguelov, Dumitru Erhan, Christian Szegedy, Scott~E. Reed,
  Cheng{-}Yang Fu, and Alexander~C. Berg.
\newblock {SSD:} single shot multibox detector.
\newblock In {\em {ECCV}}, 2016.

\bibitem{loshchilov2016sgdr}
Ilya Loshchilov and Frank Hutter.
\newblock Sgdr: Stochastic gradient descent with warm restarts.
\newblock {\em arXiv:1608.03983}, 2016.

\bibitem{LuoLUZ16}
Wenjie Luo, Yujia Li, Raquel Urtasun, and Richard~S. Zemel.
\newblock Understanding the effective receptive field in deep convolutional
  neural networks.
\newblock In {\em NIPS}, 2016.

\bibitem{MaZZS18}
Ningning Ma, Xiangyu Zhang, Hai{-}Tao Zheng, and Jian Sun.
\newblock Shufflenet {V2:} practical guidelines for efficient {CNN}
  architecture design.
\newblock In {\em {ECCV}}, 2018.

\bibitem{NohBLSK19}
Junhyug Noh, Wonho Bae, Wonhee Lee, Jinhwan Seo, and Gunhee Kim.
\newblock Better to follow, follow to be better: Towards precise supervision of
  feature super-resolution for small object detection.
\newblock In {\em ICCV}, 2019.

\bibitem{paszke2019pytorch}
Adam Paszke, Sam Gross, Francisco Massa, Adam Lerer, James Bradbury, Gregory
  Chanan, Trevor Killeen, Zeming Lin, Natalia Gimelshein, Luca Antiga, et~al.
\newblock Pytorch: An imperative style, high-performance deep learning library.
\newblock In {\em NIPS}, 2019.

\bibitem{PengXLJZJYS18}
Chao Peng, Tete Xiao, Zeming Li, Yuning Jiang, Xiangyu Zhang, Kai Jia, Gang Yu,
  and Jian Sun.
\newblock Megdet: {A} large mini-batch object detector.
\newblock In {\em {CVPR}}, 2018.

\bibitem{QinLZBYPS19}
Zheng Qin, Zeming Li, Zhaoning Zhang, Yiping Bao, Gang Yu, Yuxing Peng, and
  Jian Sun.
\newblock Thundernet: Towards real-time generic object detection on mobile
  devices.
\newblock In {\em {ICCV}}, 2019.

\bibitem{RedmonF17}
Joseph Redmon and Ali Farhadi.
\newblock {YOLO9000:} better, faster, stronger.
\newblock In {\em {CVPR}}, 2017.

\bibitem{RedmonF18}
Joseph Redmon and Ali Farhadi.
\newblock Yolov3: An incremental improvement.
\newblock {\em arXiv:1804.02767}, 2018.

\bibitem{RenHGS15}
Shaoqing Ren, Kaiming He, Ross~B. Girshick, and Jian Sun.
\newblock Faster {R-CNN:} towards real-time object detection with region
  proposal networks.
\newblock In {\em {NIPS}}, 2015.

\bibitem{SandlerHZZC18}
Mark Sandler, Andrew~G. Howard, Menglong Zhu, Andrey Zhmoginov, and
  Liang{-}Chieh Chen.
\newblock Mobilenetv2: Inverted residuals and linear bottlenecks.
\newblock In {\em {CVPR}}, 2018.

\bibitem{ShrivastavaGG16}
Abhinav Shrivastava, Abhinav Gupta, and Ross~B. Girshick.
\newblock Training region-based object detectors with online hard example
  mining.
\newblock In {\em {CVPR}}, 2016.

\bibitem{TanL19}
Mingxing Tan and Quoc~V. Le.
\newblock Efficientnet: Rethinking model scaling for convolutional neural
  networks.
\newblock In {\em {ICML}}, 2019.

\bibitem{TanPL19}
Mingxing Tan, Ruoming Pang, and Quoc~V. Le.
\newblock Efficientdet: Scalable and efficient object detection.
\newblock {\em arXiv:1911.09070}, 2019.

\bibitem{PRN}
Chien{-}Yao Wang, Hong{-}Yuan~Mark Liao, Ping{-}Yang Chen, and Jun{-}Wei Hsieh.
\newblock Enriching variety of layer-wise learning information by gradient
  combination.
\newblock In {\em ICCVW}, 2019.

\bibitem{EFM}
Chien{-}Yao Wang, Hong{-}Yuan~Mark Liao, Yueh{-}Hua Wu, Ping{-}Yang Chen,
  Jun{-}Wei Hsieh, and I{-}Hau Yeh.
\newblock Cspnet: {A} new backbone that can enhance learning capability of
  {CNN}.
\newblock In {\em CVPRW}, 2020.

\bibitem{WangBL18}
Jun Wang, Tanner~A. Bohn, and Charles~X. Ling.
\newblock Pelee: {A} real-time object detection system on mobile devices.
\newblock In {\em {NIPS}}, 2018.

\bibitem{wang2019deep}
Jingdong Wang, Ke Sun, Tianheng Cheng, Borui Jiang, Chaorui Deng, Yang Zhao,
  Dong Liu, Yadong Mu, Mingkui Tan, Xinggang Wang, et~al.
\newblock Deep high-resolution representation learning for visual recognition.
\newblock {\em arXiv:1908.07919}, 2019.

\bibitem{Roofline}
Samuel Williams, Andrew Waterman, and David~A. Patterson.
\newblock Roofline: an insightful visual performance model for multicore
  architectures.
\newblock {\em Commun. {ACM}}, 2009.

\bibitem{WuDZWSWTVJK19}
Bichen Wu, Xiaoliang Dai, Peizhao Zhang, Yanghan Wang, Fei Sun, Yiming Wu,
  Yuandong Tian, Peter Vajda, Yangqing Jia, and Kurt Keutzer.
\newblock Fbnet: Hardware-aware efficient convnet design via differentiable
  neural architecture search.
\newblock In {\em {CVPR}}, 2019.

\bibitem{XiangZYA18}
Wei Xiang, Dong{-}Qing Zhang, Heather Yu, and Vassilis Athitsos.
\newblock Context-aware single-shot detector.
\newblock In {\em {WACV}}, 2018.

\bibitem{ZhangZLS18}
Xiangyu Zhang, Xinyu Zhou, Mengxiao Lin, and Jian Sun.
\newblock Shufflenet: An extremely efficient convolutional neural network for
  mobile devices.
\newblock In {\em {CVPR}}, 2018.

\end{thebibliography}
}

\appendix






\newpage

\begin{appendix}
\section{Appendix}

\subsection{Overall Computation Allocation of TinyDets}

Tab.~\ref{tab:distribution}  provides the overall computation allocation of TinyDets. The backbone network contains most of the computation budget ($\sim 70\%$) while TinyFPN, TinyRPN and R-CNN are extremely computationally cheap.

\begin{table}[h]
    \centering

    \caption{Computation allocation of different parts (MFLOPs).}
    \label{tab:distribution}
    \resizebox{0.98\linewidth}{!}{
    \begin{tabular}{l|l|l|l|l|l}
    \toprule
    Detector & Total & Backbone & TinyFPN & TinyRPN & R-CNN\\
    \midrule
    \thename-S & 495 (100\%) & 347 (70\%) & 64 (13\%) & 16 (3\%) & 68 (14\%) \\
    \thename-M & 991 (100\%)  & 703 (71\%) & 155 (16\%) & 65 (7\%) & 68 (7\%) \\
    \thename-L & 2427 (100\%)  & 1797 (74\%) & 396 (16\%) & 166 (7\%) & 68 (3\%) \\
    \bottomrule
    \end{tabular}}
\end{table}

\subsection{Detailed Backbone Information}

\thename-S is within the computation constraint of 500 MFLOPs. It takes a more lightweight variant of MobileNetV3~\cite{HowardSCCCTWZPVLA19} as the backbone, termed as MobileNetV3-D. The specification of MobileNetV3-D can be found in Tab.~\ref{tab:V3-D}. To reduce computation cost, we remove the detailed information enhancement used in \thename-M and \thename-L, and shrink the expansion size of high-level layers.

\thename-M  is  within  the  computation  constraint  of  1000  MFLOPs.  It  takes a   MobileNetV3-BC  as  the  backbone (Tab.~\ref{tab:V3-BC320}). The input resolution is $320\times320$.

\thename-L is targeted at a high accuracy. It also adopts MobileNetV3-BC as the backbone but takes a larger input resolution as $512\times512$ (Tab.~\ref{tab:V3-BC512}).

\subsection{Computation Cost of TinyFPN \& TinyRPN}

In \thename-S, different from the structure shown in Fig. 4 of the main paper, we drop the highest-resolution pyramid level (stride=$4$) to reduce computation cost.
TinyFPN contains four $1\times1$ convs and four SCConvs (Tab.~\ref{tab:FPN_S}). 
TinyRPN contains four SCConvs, four convs for predicting scores and four convs for box regression ((Tab.~\ref{tab:RPN_S}).

As shown in Fig. 4 of the main paper, TinyFPN contains five $1\times1$ convs and five SCConvs (Tab.~\ref{tab:FPN_M}). 
TinyRPN contains five SCConvs, five convs for predicting scores and five convs for box regression (Tab.~\ref{tab:RPN_M}). 

TinyFPN and TinyRPN of \thename-L have the same structure with those of \thename-M, except the resolution. TinyFPN contains five $1\times1$ convs and five SCConvs (Tab.~\ref{tab:FPN_L}).
TinyRPN contains five SCConvs, five convs for predicting scores and five convs for regression (Tab.~\ref{tab:RPN_L}).

\subsection{Computation Cost of R-CNN}
\begin{figure}[!h]
\centering
\includegraphics[width=0.98\linewidth]{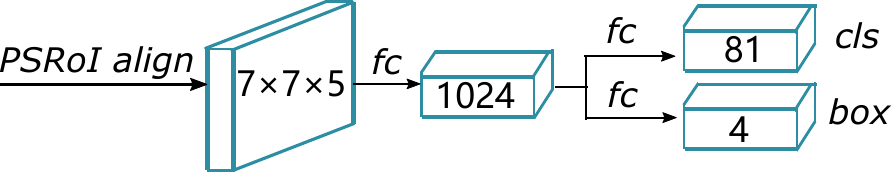}
\caption{Structure of R-CNN head.}
\label{fig:RCNN}
\vspace{1.5em}
\end{figure}

The R-CNN head is based on PSRoI align, which consists of three fully connected layers, as shown in Fig.~\ref{fig:RCNN}. 200 RoIs are used in inference phase. Based on 200 RoIs, the computation cost of R-CNN is:
\begin{equation}
    \begin{aligned}
    & 200\times (245\times1024 + 1024\times81 + 1024\times4)   \\
    =\ & 67.584 \times 10^6 \\ \approx \ & 68\ MFLOPs
    \end{aligned}
\end{equation}

\begin{table*}[htbp]
    \centering
    \caption{Specification for the backbone of \thename-S (MobileNetV3-D). Bneck denotes the inverted residual bottleneck structure~\cite{SandlerHZZC18}. ExSize denotes expansion size. SE denotes the squeeze-and-excitation module~\cite{hu2018squeeze}. NL denotes the type of the used nonlinearity. HS denotes h-swish~\cite{HowardSCCCTWZPVLA19} and RE denotes ReLU. FPN denotes whether the output of the block is fed into FPN.}
    \label{tab:V3-D}
    \resizebox{0.7\linewidth}{!}{
    \small
    \begin{tabular}{c|c|c|c|c|c|c|c|c}
    \toprule
    Input & Operator & ExSize & Out & SE & NL & Stride & FPN & MFLOPs \\
    \midrule
    $320^2\times3$ & conv2d, $3\times3$ & - & 16 & - & HS & 2 & - & 11.06 \\
    $160^2\times16$ & bneck, $3\times3$ & 16 & 16 & - & RE & 1 & - & 10.24 \\
    $160^2\times16$ & bneck, $3\times3$ & 64 & 24 & - & RE & 2 & - & 39.73\\    
    $80^2\times24$ & bneck, $3\times3$ & 72 & 24 & - & RE & 1 & - & 26.27 \\

    $80^2\times24$ & bneck, $5\times5$ & 72 & 40 & \checkmark & RE & 2 & - & 18.55 \\    
    $40^2\times40$ & bneck, $5\times5$ & 120 & 40 & \checkmark & RE & 1 & - & 20.17\\
    $40^2\times40$ & bneck, $5\times5$ & 120 & 40 & \checkmark & RE & 1 & \checkmark & 20.17 \\    

    $40^2\times40$ & bneck, $3\times3$ & 240 & 80 & - & HS & 2 & - & 23.90 \\    
    $20^2\times80$ & bneck, $3\times3$ & 240 & 80 & - & HS & 1 & - & 16.22\\
    $20^2\times80$ & bneck, $5\times5$ & 240 & 80 & - & HS & 1 & - & 17.76\\    
    $20^2\times80$ & bneck, $5\times5$ & 240 & 80 & - & HS & 1 & - & 17.76\\    

    $20^2\times80$ & bneck, $3\times3$ & 240 & 112 & \checkmark & HS & 1 & - & 19.33 \\    
    $20^2\times112$ & bneck, $3\times3$ & 336 & 112 & \checkmark & HS & 1 & \checkmark & 31.37\\    

    $20^2\times112$ & bneck, $5\times5$ & 336 & 160 & \checkmark & HS & 2 & -  & 21.33\\
    $10^2\times160$ & bneck, $5\times5$ & 480 & 160 & \checkmark & HS & 1 & -  & 16.68\\    
    $10^2\times160$ & bneck, $5\times5$ & 480 & 160 & \checkmark & HS & 1 & \checkmark  & 16.68 \\    
  
    $10^2\times160$ & bneck, $5\times5$ & 960 & 160 & \checkmark & HS
    & 2 & \checkmark & 20.26\\ 
    
    \hline
    In total &  -  & - & - & - & -
    & - & - & 347.48 \\  

    \bottomrule
    \end{tabular}
    }
\end{table*}

\begin{table*}[htbp]
    \centering
    \caption{Specification for the backbone of \thename-M. Bneck denotes the inverted residual bottleneck structure~\cite{SandlerHZZC18}. ExSize denotes expansion size. SE denotes the squeeze-and-excitation module~\cite{hu2018squeeze}. NL denotes the type of the used nonlinearity. HS denotes h-swish~\cite{HowardSCCCTWZPVLA19} and RE denotes ReLU. FPN denotes whether the output of the block is fed into FPN. Enhancement denotes modifications compared with the original MobileNetV3 and "c" denotes channel. }
    \label{tab:V3-BC320}
    \resizebox{0.7\linewidth}{!}{
    \small
    \begin{tabular}{c|c|c|c|c|c|c|c|c|c}
    \toprule
    Input & Operator & ExSize & Out & SE & NL & Stride & FPN & Enhancement & MFLOPs\\
    \midrule
    $320^2\times3$ & conv2d, $3\times3$ & - & 24 & - & HS & 2 & - & c + 50\% & 16.59\\
    $160^2\times24$ & bneck, $3\times3$ & 24 & 24 & - & RE & 1 & - & c + 50\% & 20.28\\
    $160^2\times24$ & bneck, $3\times3$ & 72 & 36 & - & RE & 2 & - & c + 50\% & 64,97\\    
    $80^2\times36$ & bneck, $3\times3$ & 108 & 36 & - & RE & 1 & - & added blocks & 55.99\\
    $80^2\times36$ & bneck, $3\times3$ & 108 & 36 & - & RE & 1 & - & added blocks & 55.99\\    
    $80^2\times36$ & bneck, $3\times3$ & 108 & 36 & - & RE & 1 & \checkmark & c + 50\% & 55.99\\
    
    $80^2\times36$ & bneck, $5\times5$ & 108 & 60 & \checkmark & RE & 2 & - & c + 50\% & 39.58\\   
    $40^2\times60$ & bneck, $5\times5$ & 180 & 60 & \checkmark & RE & 1 & - & c + 50\% & 41.78\\     
    $40^2\times60$ & bneck, $5\times5$ & 180 & 60 & \checkmark & RE & 1 & \checkmark & c + 50\% & 41.78\\    
    
    $40^2\times60$ & bneck, $3\times3$ & 240 & 80 & - & HS & 2 & - & - &31.58\\      
    $20^2\times80$ & bneck, $3\times3$ & 200 & 80 & - & HS & 1 & - & - &13.52\\  
    $20^2\times80$ & bneck, $3\times3$ & 184 & 80 & - & HS & 1 & - & - &12.44\\  
    $20^2\times80$ & bneck, $3\times3$ & 184 & 80 & - & HS & 1 & - & - &12.44\\  
    
    $20^2\times80$ & bneck, $3\times3$ & 480 & 112 & \checkmark & HS & 1 & - & - &38.71\\  
    $20^2\times112$ & bneck, $3\times3$ & 672 & 112 & \checkmark & HS & 1 & \checkmark & - & 62.86\\  
    $20^2\times112$ & bneck, $5\times5$ & 672 & 160 & \checkmark & HS & 2 & - & - &42.76\\  
    $10^2\times160$ & bneck, $5\times5$ & 960 & 160 & \checkmark & HS & 1 & - & - &33.58\\ 
    $10^2\times160$ & bneck, $5\times5$ & 960 & 160 & \checkmark & HS & 1 & \checkmark & - &33.58\\  
    
    $10^2\times160$ & bneck, $5\times5$ & 960 & 160 & \checkmark & HS & 2 & - & - & 20.26\\ 
    $5^2\times160$ & bneck, $5\times5$ & 960 & 160 & \checkmark & HS & 1 & \checkmark & - & 8.74\\     
        
    \hline
    In total &  -  & - & - & - & - & -
    & - & - & 703.42 \\  
    \bottomrule
    \end{tabular}
    }
\end{table*}

\begin{table*}[htbp]
    \centering
    \caption{Specification for the backbone of \thename-L. Bneck denotes the inverted residual bottleneck structure~\cite{SandlerHZZC18}. ExSize denotes expansion size. SE denotes the squeeze-and-excitation module~\cite{hu2018squeeze}. NL denotes the type of the used nonlinearity. HS denotes h-swish~\cite{HowardSCCCTWZPVLA19} and RE denotes ReLU. FPN denotes whether the output of the block is fed into FPN. Enhancement denotes modifications compared with the original MobileNetV3 and "c" denotes channel. }
    \label{tab:V3-BC512}
    \resizebox{0.7\linewidth}{!}{
    \small
    \begin{tabular}{l|l|c|c|c|c|c|c|c|c}
    \toprule
    Input & Operator & ExSize & Out & SE & NL & Stride & FPN & Enhancement & MFLOPs\\
    \midrule
    $512^2\times3$ & conv2d, $3\times3$ & - & 24 & - & HS & 2 & - & c + 50\% &42.47\\
    $256^2\times24$ & bneck, $3\times3$ & 24 & 24 & - & RE & 1 & - & c + 50\% &51.91\\
    $256^2\times24$ & bneck, $3\times3$ & 72 & 36 & - & RE & 2 & - & c + 50\% &166.33\\    
    $128^2\times36$ & bneck, $3\times3$ & 108 & 36 & - & RE & 1 & - & added blocks &143.33\\
    $128^2\times36$ & bneck, $3\times3$ & 108 & 36 & - & RE & 1 & - & added blocks & 143.33\\    
    $128^2\times36$ & bneck, $3\times3$ & 108 & 36 & - & RE & 1 & \checkmark & c + 50\% &143.33\\
    
    $128^2\times36$ & bneck, $5\times5$ & 108 & 60 & \checkmark & RE & 2 & - & c + 50\% &101.31\\   
    $64^2\times60$ & bneck, $5\times5$ & 180 & 60 & \checkmark & RE & 1 & - & c + 50\% &106.92\\     
    $64^2\times60$ & bneck, $5\times5$ & 180 & 60 & \checkmark & RE & 1 & \checkmark & c + 50\% &106.92\\    
    
    $64^2\times60$ & bneck, $3\times3$ & 240 & 80 & - & HS & 2 & - & - &80.86\\      
    $32^2\times80$ & bneck, $3\times3$ & 200 & 80 & - & HS & 1 & - & - &34.61\\  
    $32^2\times80$ & bneck, $3\times3$ & 184 & 80 & - & HS & 1 & - & - &31.84\\  
    $32^2\times80$ & bneck, $3\times3$ & 184 & 80 & - & HS & 1 & - & - &31.84\\  
    
    $32^2\times80$ & bneck, $3\times3$ & 480 & 112 & \checkmark & HS & 1 & - & - &98.91\\  
    $32^2\times112$ & bneck, $3\times3$ & 672 & 112 & \checkmark & HS & 1 & \checkmark & - &160.56\\  
    $32^2\times112$ & bneck, $5\times5$ & 672 & 160 & \checkmark & HS & 2 & - & - &109.12\\  
    $16^2\times160$ & bneck, $5\times5$ & 960 & 160 & \checkmark & HS & 1 & - & - &85.25\\ 
    $16^2\times160$ & bneck, $5\times5$ & 960 & 160 & \checkmark & HS & 1 & \checkmark & - &85.25\\  
    
    $16^2\times160$ & bneck, $5\times5$ & 960 & 160 & \checkmark & HS & 2 & - & - &51.15\\ 
    $8^2\times160$ & bneck, $5\times5$ & 960 & 160 & \checkmark & HS & 1 & \checkmark & - &21.66\\     
        
    \hline
    In total &  -  & - & - & - & - & -
    & - & - & 1796.90 \\  
    \bottomrule
    \end{tabular}
    }
\end{table*}

\begin{table*}[!h]
    \centering
    \begin{minipage}[c]{0.45\linewidth}
    \vspace{-35pt}
    \caption{Computation cost of TinyFPN  of \thename-S. }
    \label{tab:FPN_S}
    \resizebox{\linewidth}{!}{
    \small
    \begin{tabular}{l|c|c|c}
    \toprule
    Operator & Input  & Output  & MFLOPs\\
    \midrule
    Conv, $1\times1$ & $40^2\times40$ &  $40^2\times245$ & 16.07\\
    Conv, $1\times1$ & $20^2\times112$ &  $20^2\times245$ & 11.01\\
    Conv, $1\times1$ & $10^2\times160$ & $10^2\times245$ & 3.95\\
    Conv, $1\times1$ & $5^2\times160$ & $5^2\times245$ & 0.99\\

    SCConv, g=7 & $40^2\times245$ & $40^2\times245$ & 18.03\\
    SCConv, g=5 & $20^2\times245$ & $20^2\times245$ & 5.88\\
    SCConv, g=1 & $10^2\times245$ & $10^2\times245$ & 6.27\\
    SCConv, g=1 & $5^2\times245$ & $5^2\times245$ & 1.57\\
    \hline
    In total & - & - & 63.77\\ 
    \bottomrule
    \end{tabular}}
    \end{minipage}
    \begin{minipage}[c]{0.45\linewidth}
    \caption{Computation cost of TinyRPN  of \thename-S. }
    \label{tab:RPN_S}
    \resizebox{\linewidth}{!}{
    \small
    \begin{tabular}{l|c|c|c}
    \toprule
    Operator & Input  & Output  & MFLOPs\\
    \midrule
    \multirow{4}{*}{SCConv, g=49} 
    & $40^2\times245$ & $40^2\times245$ 
    & \multirow{4}{*}{8.33}\\ 
    & $20^2\times245$ & $20^2\times245$ \\ 
    & $10^2\times245$ & $10^2\times245$ \\ 
    & $5^2\times245$ & $5^2\times245$ \\
    \hline
    \multirow{4}{*}{Conv, $1\times1$ (score)} 
    & $40^2\times245$ & $40^2\times3$ 
    & \multirow{4}{*}{1.57}\\ 
    & $20^2\times245$ & $20^2\times3$ \\ 
    & $10^2\times245$ & $10^2\times3$ \\ 
    & $5^2\times245$ & $5^2\times3$ \\
    \hline
    \multirow{4}{*}{Conv, $1\times1$ (box)} 
    & $40^2\times245$ & $40^2\times12$ 
    & \multirow{4}{*}{6.27}\\ 
    & $20^2\times245$ & $20^2\times12$ \\ 
    & $10^2\times245$ & $10^2\times12$ \\ 
    & $5^2\times245$ & $5^2\times12$ \\
    \hline
    In total & - & - & 16.17\\
    \bottomrule
    \end{tabular}}
    \end{minipage}
\end{table*}

\begin{table*}[!h]
    \centering
    \begin{minipage}[c]{0.45\linewidth}
    \vspace{-50pt}

    \caption{Computation cost of TinyFPN  of \thename-M. }
    \label{tab:FPN_M}
    \resizebox{\linewidth}{!}{
    \small
    \begin{tabular}{l|c|c|c}
    \toprule
    Operator & Input  & Output  & MFLOPs\\
    \midrule
    Conv, $1\times1$ & $80^2\times36$ &  $80^2\times245$ & 58.02\\
    Conv, $1\times1$ & $40^2\times60$ &  $40^2\times245$ & 23.91\\
    Conv, $1\times1$ & $20^2\times112$ &  $20^2\times245$ & 11.07\\
    Conv, $1\times1$ & $10^2\times160$ & $10^2\times245$ & 3.95\\
    Conv, $1\times1$ & $5^2\times160$ & $5^2\times245$ & 0.99\\

    SCConv, g=49 & $80^2\times245$ & $80^2\times245$ & 25.09\\
    SCConv, g=7 & $40^2\times245$ & $40^2\times245$ & 18.03\\
    SCConv, g=5 & $20^2\times245$ & $20^2\times245$ & 5.88\\
    SCConv, g=1 & $10^2\times245$ & $10^2\times245$ & 6.27\\
    SCConv, g=1 & $5^2\times245$ & $5^2\times245$ & 
    1.57\\
    \hline
    In total & - & - & 154.78\\ 
    \bottomrule
    \end{tabular}}
    \end{minipage}
    \begin{minipage}[c]{0.45\linewidth}
    \caption{Computation cost of TinyRPN  of \thename-M. }
    \label{tab:RPN_M}

    \resizebox{\linewidth}{!}{
    \small
    \begin{tabular}{l|c|c|c}
    \toprule
    Operator & Input  & Output  & MFLOPs\\
    \midrule
    \multirow{5}{*}{SCConv, g=49} 
    & $80^2\times245$ & $80^2\times245$ 
    & \multirow{5}{*}{33.42}\\ 
    & $40^2\times245$ & $40^2\times245$ \\
    & $20^2\times245$ & $20^2\times245$ \\ 
    & $10^2\times245$ & $10^2\times245$ \\ 
    & $5^2\times245$ & $5^2\times245$ \\
    \hline
    \multirow{5}{*}{Conv, $1\times1$ (score)} 
    & $80^2\times245$ & $80^2\times3$ 
    & \multirow{5}{*}{6.29}\\ 
    & $40^2\times245$ & $40^2\times3$ \\ 
    & $20^2\times245$ & $20^2\times3$ \\ 
    & $10^2\times245$ & $10^2\times3$ \\ 
    & $5^2\times245$ & $5^2\times3$ \\
    \hline
    \multirow{5}{*}{Conv, $1\times1$ (box)} 
    & $80^2\times245$ & $80^2\times12$ 
    & \multirow{5}{*}{25.17}\\ 
    & $40^2\times245$ & $40^2\times12$ \\ 
    & $20^2\times245$ & $20^2\times12$ \\ 
    & $10^2\times245$ & $10^2\times12$ \\ 
    & $5^2\times245$ & $5^2\times12$ \\
    \hline
    In total & - & - & 64.88\\
    \bottomrule
    \end{tabular}}
    \end{minipage}
    
\end{table*}

\begin{table*}[!h]
    \centering
    \begin{minipage}[c]{0.45\linewidth}

    \vspace{-60pt}

    \caption{Computation cost of TinyFPN  of \thename-L. }
    \label{tab:FPN_L}
    \small
    \resizebox{\linewidth}{!}{
    \begin{tabular}{l|c|c|c}
    \toprule
    Operator & Input  & Output  & MFLOPs\\
    \midrule
    Conv, $1\times1$ & $128^2\times36$ &  $128^2\times245$ & 148.52\\
    Conv, $1\times1$ & $64^2\times60$ &  $64^2\times245$ & 61.22\\
    Conv, $1\times1$ & $32^2\times112$ &  $32^2\times245$ & 28.35\\
    Conv, $1\times1$ & $16^2\times160$ & $16^2\times245$ & 10.10\\
    Conv, $1\times1$ & $8^2\times160$ & $8^2\times245$ & 2.52\\

    SCConv, g=49 & $128^2\times245$ & $128^2\times245$ & 64.23\\
    SCConv, g=7 & $64^2\times245$ & $64^2\times245$ & 46.16\\
    SCConv, g=5 & $32^2\times245$ & $32^2\times245$ & 15.05\\
    SCConv, g=1 & $16^2\times245$ & $16^2\times245$ & 16.06\\
    SCConv, g=1 & $8^2\times245$ & $8^2\times245$ & 
    4.01\\
    \hline
    In total & - & - & 396.22\\ 
    \bottomrule
    \end{tabular}}
    
    \end{minipage}
    \begin{minipage}[c]{0.5\linewidth}

    \caption{Computation cost of TinyRPN  of \thename-L. }
    \label{tab:RPN_L}
    \resizebox{\linewidth}{!}{
    \small
    \begin{tabular}{l|c|c|c}
    \toprule
    Operator & Input  & Output  & MFLOPs\\
    \midrule
    \multirow{5}{*}{SCConv, g=49} 
    & $128^2\times245$ & $128^2\times245$ 
    & \multirow{5}{*}{85.55}\\ 
    & $64^2\times245$ & $64^2\times245$ \\
    & $32^2\times245$ & $32^2\times245$ \\ 
    & $16^2\times245$ & $16^2\times245$ \\ 
    & $8^2\times245$ & $8^2\times245$ \\
    \hline
    \multirow{5}{*}{Conv, $1\times1$ (score)} 
    & $128^2\times245$ & $128^2\times3$ 
    & \multirow{5}{*}{16.11}\\ 
    & $64^2\times245$ & $64^2\times3$ \\ 
    & $32^2\times245$ & $32^2\times3$ \\ 
    & $16^2\times245$ & $16^2\times3$ \\ 
    & $8^2\times245$ & $8^2\times3$ \\
    \hline
    \multirow{5}{*}{Conv, $1\times1$ (box)} 
    & $128^2\times245$ & $128^2\times12$ 
    & \multirow{5}{*}{64.42}\\ 
    & $64^2\times245$ & $64^2\times12$ \\ 
    & $32^2\times245$ & $32^2\times12$ \\ 
    & $16^2\times245$ & $16^2\times12$ \\ 
    & $8^2\times245$ & $8^2\times12$ \\
    \hline
    In total & - & - & 166.08\\
    \bottomrule
    \end{tabular}}
    
    \end{minipage}

\end{table*}
\end{appendix}

\end{document}